\documentclass[10pt,twocolumn,letterpaper]{article}
\usepackage[table]{xcolor}

\usepackage{iccv}              

\usepackage{amsmath,amsfonts,bm}

\def\eqref#1{equation~\ref{#1}}

\def\1{\bm{1}}

\def\rd{{\textnormal{d}}}

\def\rvf{{\mathbf{f}}}

\def\rvr{{\mathbf{r}}}
\def\rvs{{\mathbf{s}}}

\def\rvx{{\mathbf{x}}}

\DeclareMathAlphabet{\mathsfit}{\encodingdefault}{\sfdefault}{m}{sl}
\SetMathAlphabet{\mathsfit}{bold}{\encodingdefault}{\sfdefault}{bx}{n}

\def\sN{{\mathbb{N}}}

\DeclareMathOperator*{\argmax}{arg\,max}

\definecolor{iccvblue}{rgb}{0.21,0.49,0.74}
\usepackage[pagebackref,breaklinks,colorlinks,allcolors=iccvblue]{hyperref}
\usepackage{algorithm}
\usepackage{algorithmic}
\usepackage{amsthm}
\usepackage{multirow}

\newcommand{\alg}{{\textsc{Chords}}}

\theoremstyle{plain}
\newtheorem{theorem}{Theorem}[section]
\newtheorem{proposition}[theorem]{Proposition}

\theoremstyle{definition}
\newtheorem{definition}[theorem]{Definition}
\newtheorem{framework}[theorem]{Framework}

\theoremstyle{remark}

\def\paperID{9107} 
\def\confName{ICCV}
\def\confYear{2025}

\title{CHORDS: Diffusion Sampling Accelerator with \\Multi-core Hierarchical ODE Solvers}

\author{Jiaqi Han$^\ast$, Haotian Ye$^\ast$, Puheng Li, Minkai Xu, James Zou, Stefano Ermon\\
Stanford University\\
{\tt\small \{jiaqihan,haotianye,puhengli,minkai,jamesz,ermon\}@stanford.edu}
}

\begin{document}

\twocolumn[{
\maketitle
\centering
\vspace{-0.5cm}
\captionsetup{type=figure}
\includegraphics[width=\linewidth]{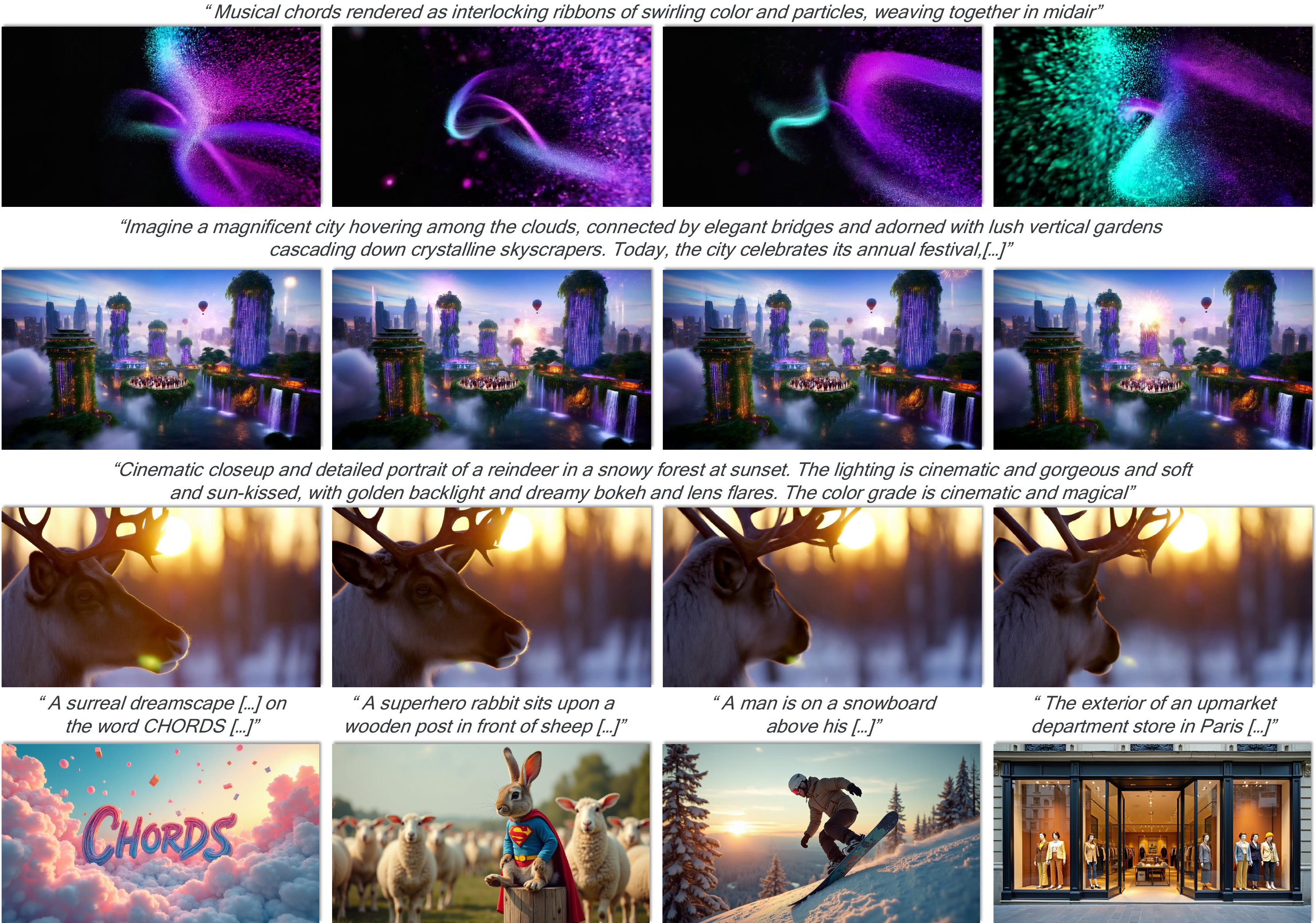}
\caption{We apply~\alg, our proposed multi-core diffusion sampling accelerator on state-of-the-art video generation model (HunyuanVideo) and image generation model (Flux). It achieves a significant $2.1\times\sim2.9\times$ speedup with four to eight computation cores.}
\label{fig:show_muscle}
\vspace{1.0cm}
}]

\def\thefootnote{*}\footnotetext{Equal contribution.} 
\def\thefootnote{\arabic{footnote}}



\begin{abstract}
Diffusion-based generative models have become dominant generators of high-fidelity images and videos but remain limited by their computationally expensive inference procedures. Existing acceleration techniques either require extensive model retraining or compromise significantly on sample quality. This paper explores a general, training-free, and model-agnostic acceleration strategy via multi-core parallelism. Our framework views multi-core diffusion sampling as an ODE solver pipeline, where slower yet accurate solvers progressively rectify faster solvers through a theoretically justified inter-core communication mechanism. This motivates our multi-core training-free diffusion sampling accelerator, \alg, which is compatible with various diffusion samplers, model architectures, and modalities. Through extensive experiments, \alg~significantly accelerates sampling across diverse large-scale image and video diffusion models, yielding up to $2.1\times$ speedup with four cores, improving by 50\% over baselines, and $2.9\times$ speedup with eight cores, all without quality degradation. This advancement enables \alg~to establish a solid foundation for real-time, high-fidelity diffusion generation. Code and demos are available at the project page:~\url{https://hanjq17.github.io/CHORDS}.
\end{abstract}


\section{Introduction}
\label{sec:intro}

Diffusion-based generative models~\cite{song2020score,ho2020denoising,sohl2015deep} have emerged as powerful tools for generating diverse and high-fidelity content across various modalities~\cite{han2024geometric,yang2023diffusionm}, including images~\cite{rombach2022high,esser2024scaling,karras2022elucidating}, videos~\cite{ho2022video,genmo2024mochi,kong2024hunyuanvideo}, and audio~\cite{kong2021diffwave,ruan2023mm,liu2023audioldm}. However, the inherently iterative sampling process makes diffusion models computationally expensive at inference time. This limitation significantly restricts their deployment in latency-sensitive or interactive scenarios, such as real-time editing~\cite{meng2021sdedit,lu2024regiondrag} and streaming applications~\cite{yin2024slow}. To address this, existing efforts have either distilled diffusion models into fewer-step samplers through additional training~\cite{salimans2022progressive,yin2024one,song2023consistency} or designed specialized sampling strategies based on ordinary differential equations (ODEs)~\cite{song2020denoising,lu2022dpm,lu2022dpmpp,zhang2022fast}. However, due to the intrinsic complexity of data, the potential for further efficiency improvement remains limited, unless output quality is substantially sacrificed~\cite{selvam2024selfrefining}.

Inspired by traditional ODE acceleration algorithms that speed up ODE solvers by incorporating multiple computational resources~\cite{shih2024parallel,junkins2013picard}, this paper explores training-free diffusion acceleration \textit{via parallel computation cores}. Specifically, we seek algorithms that (1) scale seamlessly with variable computational cores—particularly multiple GPUs or TPUs—to flexibly speed up generation; (2) remain universally applicable to arbitrary diffusion-based generative models without restrictions on model architectures, noise schedules, or training procedures; and (3) require no additional training, maintaining broad deployability and ease of adoption. Unfortunately, existing multi-core acceleration methods are either architecture-specific~\cite{fang2024xdit}, training-based~\cite{song2023consistency,kim2023consistency}, or resource-constrained~\cite{selvam2024selfrefining}, thus falling short of these requirements.

In practice, designing principled algorithms that achieve speed-up with multi-core without harming output quality is highly non-trivial, especially given the complexity of neural-network-based diffusion models \cite{song2020denoising}. To tackle this problem, we propose a novel and general algorithmic framework for multi-core diffusion sampling acceleration. This framework is built upon a \textit{inter-core rectification technique}, which enhances a fast solver by leveraging information from a slower but more accurate solver, and thus provides theoretical guarantees for improved efficiency. By iteratively applying this technique across a chain of hierarchical solvers from slow to fast, we can efficiently propagate information across computation cores and substantially enhance efficiency without compromising generation quality. 
The proposed framework encompasses a family of acceleration algorithms, where different instantiations of algorithms are realized by different sequences that specify the speedup ratio between each core pair.
We show several existing methods~\cite{shih2024parallel, selvam2024selfrefining} as concrete instantiations of our proposed framework, and provide theoretical justification for the optimal parameter within the framework. 

By leveraging results obtained from our framework, we propose \alg, a multi-\textbf{C}ore, \textbf{H}ierarchical \textbf{O}DE-based \textbf{R}ectification for \textbf{D}iffusion \textbf{S}ampling acceleration. As a training-free, model-agnostic method, \alg~can be readily adapted to various diffusion solvers for both image and video generation, achieving significant acceleration with negligible quality degradation under arbitrary numbers of computational cores. 
We empirically benchmark \alg~against existing parallel diffusion samplers on five state-of-the-art video and image diffusion models, and observe that our approach can consistently reduce sampling latency without measurable quality degradation. Compared with vanilla sequential sampling using 50 inference steps, \alg~achieves a $2.1\times \sim 2.9\times$ speedup using $4\sim 8$ GPU cores on 
HunyuanVideo~\cite{kong2024hunyuanvideo}, and similarly for other models. It outperforms previous SOTA acceleration sampling~\cite{selvam2024selfrefining} algorithms by up to 50\% with eight cores and 67\% with four cores, while offering more flexibility, adaptivity, and robustness.

We summarize our major contributions below:
\begin{itemize}
    \item We propose a novel framework for training-free multi-core diffusion acceleration. Within the framework, we systematically study the hierarchical inter-core rectification technique that leverages parallel computational resources, and justify the best practice of instantiating empirical algorithms.

    \item We develop \alg, a flexible, scalable, and model-agnostic diffusion sampling accelerator applicable to diverse diffusion models. \alg~take advantage of multiple computation cores and initializes a hierarchy of slow-to-fast solvers, and provides efficient yet accurate outputs via information propagation across solvers.

    \item We demonstrate through extensive benchmarking on state-of-the-art diffusion models that \alg~consistently achieves significant sampling latency reductions (up to $2.9\times$ speedup) without measurable quality degradation, substantially outperforming existing methods.
\end{itemize}

\section{\alg: A Multi-core Diffusion Solver}

Given a data sampled from the data space $\mathbf x \in \mathcal X$, diffusion models feature a noising process that gradually perturbs the input towards Gaussian noise, while a neural network 
$
\bm\epsilon_\theta : \mathcal X \times \mathcal T \mapsto \mathcal X
$
is trained to approximate the score function $\nabla_\rvx\log p_t(\rvx)$ through denoising score matching~\cite{song2020score,song2019generative}, where $\mathcal T = [0, 1]$. The reverse sampling process transits from noise to data along the following \emph{probability-flow ODE} (PF-ODE):
\begin{align}
\label{eq:pf-ode}
    \rd\rvx=\left(f(t)\rvx-\frac{1}{2}g^2(t)\bm \epsilon_\theta(\rvx, t)\right)\rd t,
\end{align}
where $f(t)$ and $g(t)$ are coefficients defined by the noise schedule. 
To ensure our methodology is compatible with different parameterizations of $f(t)$ and $g(t)$ under different algorithms~\citep{ho2020denoising, song2020denoising, lipman2023flow,liu2023flow}, we consider the initial value problem as a more general form of \Cref{eq:pf-ode}:
\begin{align}
\label{eq:ode}
    \rd\rvx=\rvf_\theta\left(\rvx,t\right)\rd t,\quad\quad \rvx|_{t=0}=\rvx_0,
\end{align}
where $\rvf_\theta(\rvx, t)=f(t)\rvx-\frac{1}{2}g^2(t)\bm\epsilon_\theta(\rvx,t)$, and $\rvx_0$ is the initial condition\footnote{Since we focus on diffusion sampling, we start from $t=0$ as noise and solve toward $t=1$, different from the convention where $t=0$ refers to the data distribution.}. 

To solve \Cref{eq:ode}, empirical solvers such as DDIM~\cite{song2020denoising}, Euler~\cite{lipman2023flow}, and their high-order counterparts~\cite{lu2022dpm,zhang2022fast} need to discretize $\mathcal T$ into a sequence of time steps. 
Since the purpose of this section is to formulate the multi-core acceleration framework, we consider the \textit{continuous} setting where an ODE solver solves \Cref{eq:ode} with an indefinitely small time step. 
Throughout the section, we use terms ``solver'' and ``core'' interchangeably.
We discuss how to design empirical algorithms of the framework using discretized timesteps in \Cref{sec:alg}.

\subsection{Multi-core Rectification}
\label{sec:rectification}

The key idea of multi-core ODE solvers is to leverage the additional computation resources into speed advantage~\cite{shih2024parallel, selvam2024selfrefining}. 
This section proposes a fundamental operation, \emph{multi-core rectification}, on top of which we build up algorithms to take advantage of multiple cores effectively.

\begin{figure}[t!]
    \centering
    \includegraphics[width=0.98\linewidth]{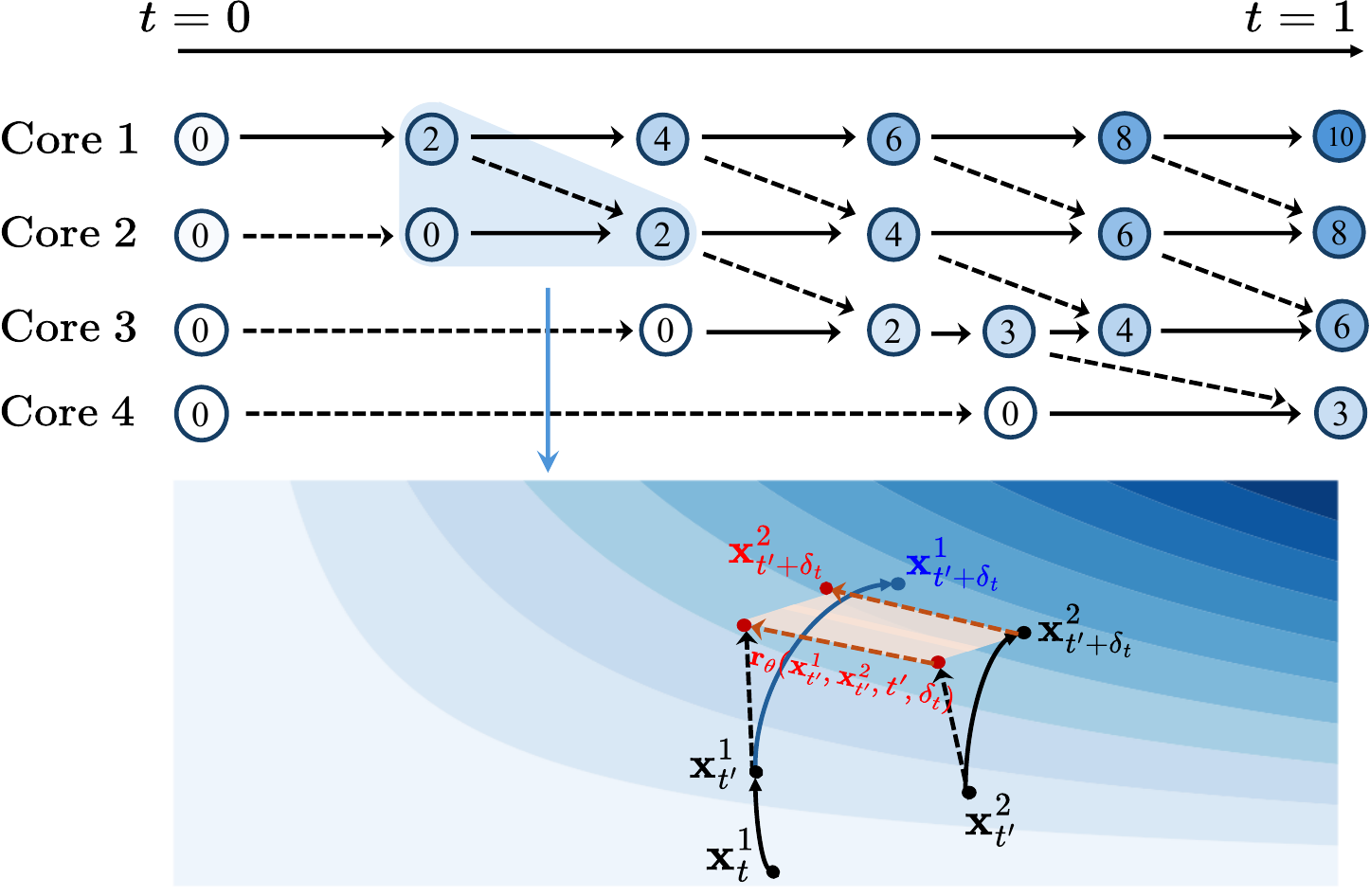}
    \caption{An illustration of Framework~\ref{def:alg} with $K=4$ cores. According to \Cref{sec:selection_sequence}, four cores are initialized at time $\mathbf I = [0, 0.2, 0.4, 0.7]$. 
    Then, each core solves forward simultaneously, where the number in marks represents the wall-clock time of that solver (multiplied by $10$). Solid lines denote the continuous solve within a core, and dash lines across cores denote the rectification technique proposed in \Cref{sec:rectification}. The bottom figure illustrates the update of $\rvx^2_{t'+\delta_t} $ with $t = 0.2, t'=0.4$. Notice that \textcolor{red}{$\rvx^2_{t'+\delta_t}$} is closer to $\textcolor{blue}{\rvx^1_{t'+\delta_t}}$ than $\rvx^2_{t'+\delta_t}$, the latent before rectification.
    }
    \label{fig:alg_framework}
    \vspace{-5pt}
\end{figure}

\looseness=-1
Consider a simple case with two cores (solvers) starting from different time $t < t'$ with a prior guess of $\rvx_{t}^1, \rvx^2_{t'}$ (initialization of guess will be specified in \Cref{sec:selection_sequence}). 
After time $\delta_t=t'-t$, the slow solver $1$ will solve $\rvx^1_{t}$ to $\rvx^1_{t'}$ (which is more accurate than $\rvx^2_{t'}$), and the fast solver $2$ will reach $\rvx^2_{t'+\delta_t}$. The following update rule leverages $\rvx^1_{t'}$ for rectifying the fast solver that has already solved forward to ${\rvx}_{t'+\delta_t}$:
\begin{align}
\label{eq:update_basis}
    &{{\mathbf x}_{t' +\delta_t}^2 }\leftarrow  {\mathbf x}_{t'+\delta_t}^2 + \mathbf r_\theta ( \rvx^1_{t'},  \rvx^2_{t'}, t', \delta_t), \quad\text{where }\\
    &  \mathbf r_\theta ( \rvx_t,  \tilde \rvx_t, t, \delta_t) \triangleq \delta_t \cdot \big(\mathbf f_\theta (\rvx _{t}, t) - \rvf_\theta  (\tilde \rvx_t, t)\big) + \rvx_t - \tilde \rvx_t. 
\end{align} 
This update rule is a widely adopted update mechanism in traditional multi-grid ODE solvers~\cite{brandt1977multi,greenbaum1986multigrid}, as illustrated in \Cref{fig:alg_framework}. 
The rectification term can be regarded as performing two ``one-step jumps'' from $t'$ to $t'+\delta_t$ as estimation of continuous solves, and computing the difference between two estimated terms to refine the continuous solve from the less accurate $\rvx^2_{t'}$.
The validity of the proposed rectification is theoretically guaranteed in the proposition below.

\begin{proposition}
\label{lemma:rectification} 
    Suppose $\mathbf f(\rvx_t , t)$ is sufficiently smooth around $(\rvx_t, t)$. For any $t<t'$, any $\tilde \rvx_t$ as an approximation of $\rvx_t$, if $\rvx_{t'}, \tilde \rvx_{t'}$ are separately solved from $\rvx_t, \tilde \rvx_t$, then the rectification reduces the approximation error, i.e.,
    \begin{align*}
        \|  \tilde \rvx_{t'} + \mathbf r_\theta (\rvx_t, \tilde \rvx_t, t, t'-t) - \rvx_{t'} \|_2  
        =o( \| \tilde \rvx_{t'}  - \rvx_{t'}  \|_2 ),
    \end{align*}
    where $o(\cdot)$ is w.r.t. to $t'-t$.
\end{proposition}

Proposition~\ref{lemma:rectification} shows that the rectification rule is beneficial: after the update, the approximation error becomes a negligible term compared with that without the update, demonstrating that the update can eliminate the error caused by the inaccuracy of $\tilde \rvx^k_{t}$. 
By further composing a hierarchy of the technique, we can ensure that fast solvers generate outputs with precision similar to slow solvers, making it a core operation for multi-core acceleration.

\subsection{Our Framework}
This paper targets an algorithmic framework that is (1) training-free, i.e. it does not require any type of model re-training; (2) model-agnostic, i.e., it works for different diffusion samplers with different model architectures and noise schedulers; and (3) robust to resources, i.e., it applies to different amount of compute cores. 
The algorithmic framework proposed below meets these requirements and assembles a family of parallel algorithms parameterized by the initialization sequence. 
Remarkably, various previous algorithms, such as \cite{selvam2024selfrefining,shih2024parallel}, can be regarded as special instantiations of $\mathcal A$ after discretization (more discussions in \Cref{sec:discuss}).
Throughout the paper, the superscripts refer to the index of cores, and the subscripts represent time.

\begin{framework}[Diffusion Solver with Multi-core Rectification]\label{def:alg}
Given any core number $K$, the parallel diffusion solver framework $\mathcal A$ consists the following components:
    \begin{enumerate}
        \item Parameterization: An algorithm is parameterized by an increasing \emph{initialization sequence} $\mathbf I = [t^{(1)}, \cdots, t^{(K)}]$, where $ 0=t^{(1)} < t^{(2)} < \cdots < t^{(K)} < 1.$
        \item Initialization: Core $k$ is initialized at time $t^{(k)}$ with the latent $\rvx ^{k}_{t^{(k)}} = \rvx_0 + t^{(k)} \rvf_\theta  (\rvx_0, 0)$. All cores simultaneously solve via the ODE $\rd \rvx = \rvf_\theta(\rvx, t)\rd t$.
        \item Termination: Core $k$ is terminated when it reaches $t=1$ with time $1 - t^{(k)}$. The algorithm terminates whenever the output $\rvx^{k}_1$ from core $k$ meets user-defined criteria. 
        \item Communication: For every duration of $\delta^{(k)} = t^{(k)} - t^{(k-1)}$, all cores with $k > 1$ that have not terminated will have their latents $\rvx^k_{t^{(k)} + n \delta^{(k)}}$ rectified with $\rvr_\theta (\rvx^{k-1}_{t^{(k-1)} + n \delta^{(k)}}, \rvx^{k}_{t^{(k-1)} + n \delta^{(k)}}, t^{(k-1)} + n \delta^{(k)}, \delta^{(k)} )$, according to \Cref{eq:update_basis}, where $n\in\sN^+$.
    \end{enumerate}
\end{framework}

Several remarks are made below. First, the framework initializes core $k$ at time $t^{(k)}$, establishing a \textit{hierarchy} of solvers from slow to fast. This design follows the intuition that slower solvers can propagate accurate solutions to faster solvers to avoid error accumulation.  
Second, the multi-core rectification is \emph{fully pipelined}: Updates are triggered whenever a slower core ``reaches'' a faster core, and rectification happens among cores hierarchically--core $k-1$ can update core $k$, while itself is updated by the preceding core. This structure ensures information flows efficiently without intermediate bubbles in the multi-core pipeline, in contrast with previous works~\cite{selvam2024selfrefining} that requires deliberate and ad-hoc design of the pipeline to maximize speedup. 
Lastly, the arrival time of solver $k$ is $1 - t^{(k)}$, with later solvers providing increasingly accurate outputs. This streaming structure can be terminated on the fly when the output quality meets user-defined criteria, such as convergence of the residual~\cite{tang2024acceleratingparallelsamplingdiffusion}. To safeguard high-quality outputs even in cases where acceleration is limited, we fix the slowest solver to initialize at $t^{(1)} = 0$.  

Framework~\ref{def:alg} yields a parallel training-free diffusion sampling accelerator which is naturally compatible with different architectures. Moreover, by parameterizing through the initialization sequence $\mathbf I$, our framework facilitates systematic analysis of the efficiency-accuracy trade-off across different settings, as we will detail in~\Cref{sec:selection_sequence}.

\subsection{Initialization Sequence Selection} 
\label{sec:selection_sequence}

The remaining question is to determine the optimal initialization sequence $\mathbf I$. We take a greedy approach--given a desired speedup ratio $s$, the fastest core should be initialized accordingly, and we aim to allocate the rest of the cores appropriately such that the eventual error of $\rvx^K_1$ is minimized. 
Due to the high dimensionality and complexity of $\rvf_\theta$, pre-computing this error is infeasible. To address this challenge, we propose to construct a \textit{surrogate function}, referred to as a ``reward function'', that captures the problem's characteristics. This function should be simple yet reasonable, supervising the selection of $\mathbf I$ and inspiring an initialization that performs effectively in practice.  

\begin{definition}[Speedup] For any initialization sequence $\mathbf I$, the speedup ratio of $\mathbf I$ is defined as $\mathcal S(\mathbf I) = \frac 1 {1 - t^{(K)}}$, i.e. how many times of acceleration that the first output can provide.
\end{definition}

\begin{definition}[Reward]\label{def:reward}
A function $\mathcal R: \mathbf I \mapsto [0,1]$ is a reward function if it satisfies that 
    \begin{itemize}
        \item (Optimality) The single-core solve has reward $\mathcal R([0]) = 1$ and $\mathcal S([0]) = 1$. For any $\mathbf I$ s.t. $\mathcal S(\mathbf I)> 1, \mathcal R(\mathbf I) \in (0, 1).$ 
        \item (Monotonicity) For any $\mathbf I_1 \subsetneq \mathbf I_2$ s.t. $\mathcal S(\mathbf I_1) = \mathcal S(\mathbf I_2)$, we have $\mathcal R(\mathbf I_1) < \mathcal R(\mathbf I_2).$ For any $\mathbf I_1$ that is a prefix of $\mathbf I_2$, we have $\mathcal R(\mathbf I_1) \geq \mathcal R(\mathbf I_2).$
        \item (Trade-off) For any speedup ratios $1 \leq s_1 < s_2$, we have $$\max_{\mathcal S(\mathbf I) = s_1} \mathcal R(\mathbf I) > \max_{\mathcal S(\mathbf I) = s_2} \mathcal R(\mathbf I).$$
    \end{itemize}
    In this paper, for simplicity we select $\mathcal R(\mathbf I) = \sum_{d=1}^D \ln x_{1,d}^K $ when $\rvf_\theta(\rvx, t) = \rvx $ and $\rvx_0 = \bm{1}$ with $\rvx_1^K=(x_{1,1}^K,\cdots,x_{1,D}^K)$ being the final latent of the fastest core and $D$ being the latent dimension. The proof of satisfaction is in the appendix.
\end{definition}



Definition~\ref{def:reward} highlights crucial properties that make the reward surrogate simple, reasonable, and effective. The monotonicity ensures that adding cores in the middle never worsens the reward, and the streaming outputs have increasing rewards. The trade-off property shows that to gain higher speedups, we inevitably sacrifice the reward. These properties closely mirror the practical reality of parallelizing ODE solvers for diffusion sampling, thereby justifying $\mathcal R$ as a good indicator for more complex real-world metrics.

With the proposed reward function, we seek to devise the optimal initialization $\mathbf I$ that maximizes $\mathcal R(\mathbf I)$. However, this is still challenging for generally large $K$ even for simple rewards. 
To circumvent this issue, we propose to decompose the  problem into a sequence of \textit{triple-core} optimization problems. Specifically, the slowest and fastest cores are fixed based on the speedup ratio $s$, and the intermediate cores are determined fast-to-slow. For each core $k$ to be initialized, we consider its optimal initialization given the slowest core $1$, the faster core $k+1$, but only on the sequence of $[0, t^{(k+1)}]$ (or $[0,1]$ for core $K$). This greedy decomposition simplifies the problem by ignoring all cores $k' > k+1$, allowing us to apply the following theorem.



\begin{theorem}[Optimal Initialization]\label{thm:core_insert}
    For any speedup ratio $s \geq 2$ with $K=3$, the initialization sequence $\mathbf I$ with $\mathcal S(\mathbf I) =s $ that maximizes the reward function $\mathcal R$ is
    \begin{align}
        \argmax_{\mathcal S(\mathbf I) = s} \mathcal R(\mathbf I) = \begin{cases}
        [t^{(1)}=0, t^{(2)} = \frac{t^{(3)}}{2}, t^{(3)} = \frac {s-1} s ],\ \ s \leq 3 \\
            [t^{(1)} = 0, t^{(2)} = 2t^{(3)} - 1 , t^{(3)} = \frac {s-1} s], s > 3     
        \end{cases} .
        \label{eq:insert}
    \end{align}
\end{theorem}

\Cref{thm:core_insert} points out a general recipe of   initializing the solvers to maximize the reward when using three cores.
For general $K$, we iteratively determine $t^{(k)}$ for $k=K,\cdots, 1$ right to left (fast-to-slow) using the following recursion:
\begin{align*}
    t^{(k)} = \begin{cases}
        2t^{(k+1)} - t^{(k+2)} &\quad t^{(k+1)} > \frac {2t^{(k+2)}} 3 \\
        \frac {t^{(k+1)}} 2 &\quad \text{otherwise}
    \end{cases},
\end{align*}
with $t^{(K)} = \frac {s-1} s$ and $t^{(1)}=0$. As an illustration, in \Cref{fig:alg_framework} we have $K = 4, s=\frac {10}3$, resulting in $\mathbf I = [0, 0.2, 0.4,0.7]$.
Notice that the information from the most accurate (left-most) solver can quickly  propagate to the fastest (right-most) solver, enabling continuous rectification and significantly reducing error accumulation in the fast solver's trajectory. Consequently, even aggressively accelerated solvers can still produce high-quality outputs.

In summary, this section presents an algorithm with a consistent multi-core rectification rule that allows for the design of optimal multi-core acceleration algorithms. 
The proposed framework inspires a theoretically supported initialization sequence $\mathbf I$, and we demonstrate in \Cref{sec:exp} that this strategy is also empirically effective in speeding up diffusion sampling while maintaining high sample quality.

\section{Empirical Instantiation with Discretization}\label{sec:alg}

The previous section established a continuous view for multi-core diffusion ODE solvers. We now bridge this theory to a discrete-time implementation, resulting in a practical, parallel diffusion sampling acceleration algorithm. 

To solve \Cref{eq:ode} in practice, empirical single-core diffusion solvers typically discretize $\mathcal T$ into a finite sequence $\mathbb T = [t(0)=0, t(1), \dots, t(N)=1]\subseteq [0,1]$, where $t(\cdot)$ is the discretization function and $N$ is the number of diffusion steps. 
They initialize \(\mathbf{x}_{t(0)} = \mathbf{x}_0\) from the prior with pure Gaussian noise and iteratively compute 
\begin{align}
\mathbf{x}_{t(i+1)} \;=\; \mathbf{x}_{t(i)} \;+\;\rvs_\theta\bigl(\mathbf{x}_{t(i)}, t(i), t(i+1)\bigr),  \label{eq:solver}
\end{align}
where $\rvs_\theta$ represents a single forward step by calling the neural network $\rvf_\theta$. For instance, DDIM and Euler solver takes the form of  $\rvs_\theta(\rvx, t, t') = (t'-t) \rvf_\theta(\rvx, t)$. 
After \(N\) iterations, we obtain the final solution \(\mathbf{x}_{1}\). 
The empirical speedup ratio is measured by the number of sequential network passes since they dominate the runtime. 
Unsurprisingly, using a smaller $N$ increases speed but degrades sample quality.


\begin{algorithm}[t!]
\small
\caption{\alg: Multi-\textbf{C}ore \textbf{H}ierarchical \textbf{O}DE-based \textbf{R}ectification for \textbf{D}iffusion \textbf{S}ampler}
\label{alg:algo-main}
\textbf{Input:} Diffusion step $N$, time discretization function $t(\cdot)$, number of cores $K$, initial condition $\rvx_0$, solver $\rvs_\theta$, and refiner $\rvr_\theta$.
\begin{algorithmic}[1]
\STATE $\rvx^k\leftarrow \rvx_0,\rvx^k_\text{prev}\leftarrow \rvx_0,\forall k \in [1,...,K]$
\FOR {$\text{step} = 1,\cdots, N$}
    \STATE \textcolor{gray}{// Simultaneously for all cores $k = 1,\cdots, K$}
    \STATE $\text{cur},\text{next}\leftarrow\texttt{Scheduler}(N,\text{step},k)$    \textcolor{gray}{// \Cref{eq:scheduler}}
    \STATE $\Delta \leftarrow \rvs_\theta(\rvx^k,t(\text{cur}),t(\text{next}))$
    \STATE $\text{prev}, \_ \leftarrow \texttt{Scheduler}(N, \text{step},k-1)$    \textcolor{gray}{// \Cref{eq:scheduler}}
    \IF {$\texttt{Communicate}(k, \text{prev}, \text{cur})$}
        \STATE $\Delta \leftarrow \Delta +\rvr_\theta(\rvx^{k-1},\rvx^k_\text{prev}, t(\text{prev}),t(\text{next})-t(\text{prev}))$
        \STATE $\rvx^k_\text{prev}\leftarrow\rvx^k+\Delta$
        \ENDIF
        \STATE $\textbf{synchronize}$
        \STATE $\rvx^k\leftarrow\rvx^k+\Delta$
    \IF {$\text{next} = N$}
        \STATE Core $k$ returns output $\rvx^k$ on-the-fly
        \STATE \textbf{terminate} if user-defined criteria is satisfied
    \ENDIF
    
\ENDFOR
\end{algorithmic}
\end{algorithm}

\begin{table*}[t!]
  \centering
  \caption{Benchmark results of parallel diffusion solvers on video diffusion models using VBench. We evaluate on three video diffusion models with the number of cores $K$ set to 4, 6, and 8. Our approach achieves the highest speedup without measurable quality degradation.}
  \vskip -0.1in
    \resizebox{\linewidth}{!}{%
    \begin{tabular}{clcccccccccccc}
    \toprule
          &       & \multicolumn{4}{c}{\textbf{Num Core = 4}} & \multicolumn{4}{c}{\textbf{Num Core = 6}} & \multicolumn{4}{c}{\textbf{Num Core = 8}} \\
          \cmidrule(lr){3-6}\cmidrule(lr){7-10}\cmidrule(lr){11-14}
          &       & Time per & Speed- & VBench & Latent  & Time per & Speed- & VBench & Latent  & Time per & Speed- & VBench & Latent \\
          &       & sample $\downarrow$ & up $\uparrow$    & Quality $\uparrow$ & RMSE $\downarrow$  & sample $\downarrow$ & up $\uparrow$    & Quality $\uparrow$ & RMSE $\downarrow$  & sample $\downarrow$ & up $\uparrow$    & Quality $\uparrow$ & RMSE $\downarrow$ \\
    \midrule
    \multirow{4}[2]{*}{HunyuanVideo} & Sequential &  372.5     &   -    &  84.4\%     &   -    &   372.5    &   -    &   84.4\%    &   -    &  372.5     &   -    & 84.4\%      & - \\
          & ParaDIGMS~\cite{shih2024parallel} &   286.8    &   1.3    &   84.2\%    &    0.188   &  280.3     &  1.4     &   84.3\%    &   0.203    &   268.1    &   1.4    &   84.2\%    & 0.202 \\
          & SRDS~\cite{selvam2024selfrefining}  &   257.4   &  1.4     &  84.2\%     &   0.068    &   171.6    &   2.2   &  84.2\%     &  0.068     &  147.1     & 2.6      &  84.2\%     & 0.068 \\
          & \cellcolor{gray!15}Ours  &  \cellcolor{gray!15}{183.1}     &    \cellcolor{gray!15}{\textbf{2.1}}   &   \cellcolor{gray!15}{84.1\%}    &  \cellcolor{gray!15}{0.066}     &   \cellcolor{gray!15}{154.2}    &   \cellcolor{gray!15}{\textbf{2.5}}    &  \cellcolor{gray!15}{84.2\%}     &  \cellcolor{gray!15}{0.067}     &    \cellcolor{gray!15}{129.1}  &   \cellcolor{gray!15}{\textbf{2.9}}   &   \cellcolor{gray!15}{84.1\%}    & \cellcolor{gray!15}{0.068} \\
    \midrule
    \multirow{4}[1]{*}{Wan2.1} & Sequential &  494.5     &  -     &  85.3\%     &   -    & 494.5      &  -     &  85.3\%     &  -     &   494.5    &   -    &  85.3\%     & - \\
          & ParaDIGMS~\cite{shih2024parallel} &  312.1     &  1.5     &   85.2\%    &  0.083     &  295.6     &  1.6     &  85.1\%     &   0.090    &  288.1     &  1.7    &  85.2\%     & 0.097 \\
          & SRDS~\cite{selvam2024selfrefining}  &   375.4    &  1.3     &   85.1\%    &  0.041     &  250.3     & 2.0      &   85.1\%    &   0.043    &   214.5    &   2.3    &   85.1\%    & 0.041 \\
          & \cellcolor{gray!15}Ours  &  \cellcolor{gray!15}{288.4}     &   \cellcolor{gray!15}{\textbf{1.8}}   &   \cellcolor{gray!15}{85.1\%}    &  \cellcolor{gray!15}{0.034}     &   \cellcolor{gray!15}{228.7}    &   \cellcolor{gray!15}{\textbf{2.3}}    &  \cellcolor{gray!15}{85.1\%}     &  \cellcolor{gray!15}{0.040}     &    \cellcolor{gray!15}{189.4}   &   \cellcolor{gray!15}{\textbf{2.7}}    &   \cellcolor{gray!15}{85.1\%}    & \cellcolor{gray!15}{0.043} \\
    \midrule
    \multirow{4}[1]{*}{CogVideoX1.5} & Sequential &   474.5    &  -     &   81.9\%    &   -    &   474.5      &    -   &   81.9\%    &   -    &    474.5     &   -    &    81.9\%   &  -  \\
          & ParaDIGMS~\cite{shih2024parallel} &   365.0    &  1.3     &   81.6\%    &  0.132     &   335.1    &  1.4     & 81.6\%     &   0.159    &   291.1    &  1.6     &  81.5\%     &  0.187 \\
          & SRDS~\cite{selvam2024selfrefining}  &   386.9    &  1.2     &  81.5\%     & 0.056      &  258.0     &    1.8  &    81.5\%   &   0.056    &   221.1    &   2.1    &   81.5\%    & 0.056 \\
          & \cellcolor{gray!15}Ours  &  \cellcolor{gray!15}{256.8}     &    \cellcolor{gray!15}{\textbf{2.0}}   &   \cellcolor{gray!15}{81.8\%}    &  \cellcolor{gray!15}{0.054}     &   \cellcolor{gray!15}{221.5}    &   \cellcolor{gray!15}{\textbf{2.3}}    &  \cellcolor{gray!15}{81.7\%}     &  \cellcolor{gray!15}{0.054}     &    \cellcolor{gray!15}{202.5}   &   \cellcolor{gray!15}{\textbf{2.4}}    &   \cellcolor{gray!15}{81.8\%}    & \cellcolor{gray!15}{0.055} \\
    \bottomrule
    \end{tabular}%
    }
  \label{tab:video_overall_table}%
  \vskip -0.1in
\end{table*}%

In order to leverage our Framework~\ref{def:alg} with empirical solvers such as DDIM and Euler solver, we present a discretized version of~\alg~as a practical instantiation, with details of each components specified as follows.


\noindent \textbf{Initialization.} 
Following our framework, we need to determine at which discretized steps a core should be initialized.
Specifically, the initialization $\hat{\mathbf I }= [i_1,\cdots, i_K]$ is a subsequence of $ [0,1,\cdots, N]$ rather than of interval $[0,1]$, corresponding to diffusion time $\mathbf I = [t({i_1}), \cdots, t(i_K)]$. 
Concrete choices of $\hat {\mathbf I}$ can be determined based on \Cref{thm:core_insert} with its detailed instantiation specified in \Cref{sec:exp}.
Now, given an $\hat {\mathbf I}$, each core $k$ instead initializes $\rvx^k_{t({i_k})}$ by iterating \Cref{eq:solver} $k-1$ times to jump from $t({i_1}), t({i_2}), \cdots$ to $t({i_k})$, forming a hierarchical slow-to-fast solvers sequence with the speedup ratio becoming $\frac{1}{1-t(i_k)+(k-1)/N}$.
Notice that this differs from the continuous setting where the initialization takes ``no time'' compared with continuous solve.
Once the core $k$ has been initialized, it solves forward step-by-step along $\mathbb T$. This implies it reaches $t(N)$ after $N - i_k$ steps. This logic is reflected in the $\texttt{Scheduler}$ in \Cref{alg:algo-main}. Specifically, the $\texttt{Scheduler}(N, \text{step}, k)$ equals
\begin{align}
     \begin{cases}
        (i_\text{step}, i_{\text{step} + 1}) &\quad \text{step} < K \\
        (i_k + \text{step} - k, i_k + \text{step} - k +1) &\quad  \text{step} \geq K
    \end{cases}.\label{eq:scheduler}
\end{align}

\noindent \textbf{Communication.} Core $k - 1$ will reach $t({i_k})$ after $i_k - i_{k-1}$ steps, thus triggering a rectification via \Cref{eq:update_basis}. Once updated, core $k$ will continue its solve based on the updated $\rvx^k$ at time step $2i_k - i_{k-1}$, until the next rectification is triggered, for every $i_k - i_{k-1}$ steps. In \Cref{alg:algo-main}, $\texttt{Communicate}(k, \text{prev}, \text{cur}) $ is \texttt{True} if and only if $k>1$ and $(\text{cur} - \text{prev})$ is dividable by $i_k - i_{k-1}$.

\noindent \textbf{Remark on \Cref{alg:algo-main}.} Notice that the algorithm interactively returns generated outputs with increasing quality, where the last output is guaranteed to be identical to the output when there is no multi-core acceleration. This ensures that users can enjoy different levels of acceleration based on their preference for efficiency over quality. Importantly, the algorithm is pipelined such that no core needs to wait for previous cores, thus maximizing the information flow within the hierarchy. Finally, our technique is training-free and model-agnostic. It can be combined with other acceleration techniques such as splitting model parameters across different cores or model distillation to further improve efficiency or quality.

In summary, \Cref{alg:algo-main} provides a flexible, model-agnostic solution that can be scaled to an arbitrary number of cores and adapted to various common diffusion algorithms, with strong empirical performance justified in the experiments section below.

\begin{figure*}[t!]
    \centering
    \includegraphics[width=1.0\linewidth]{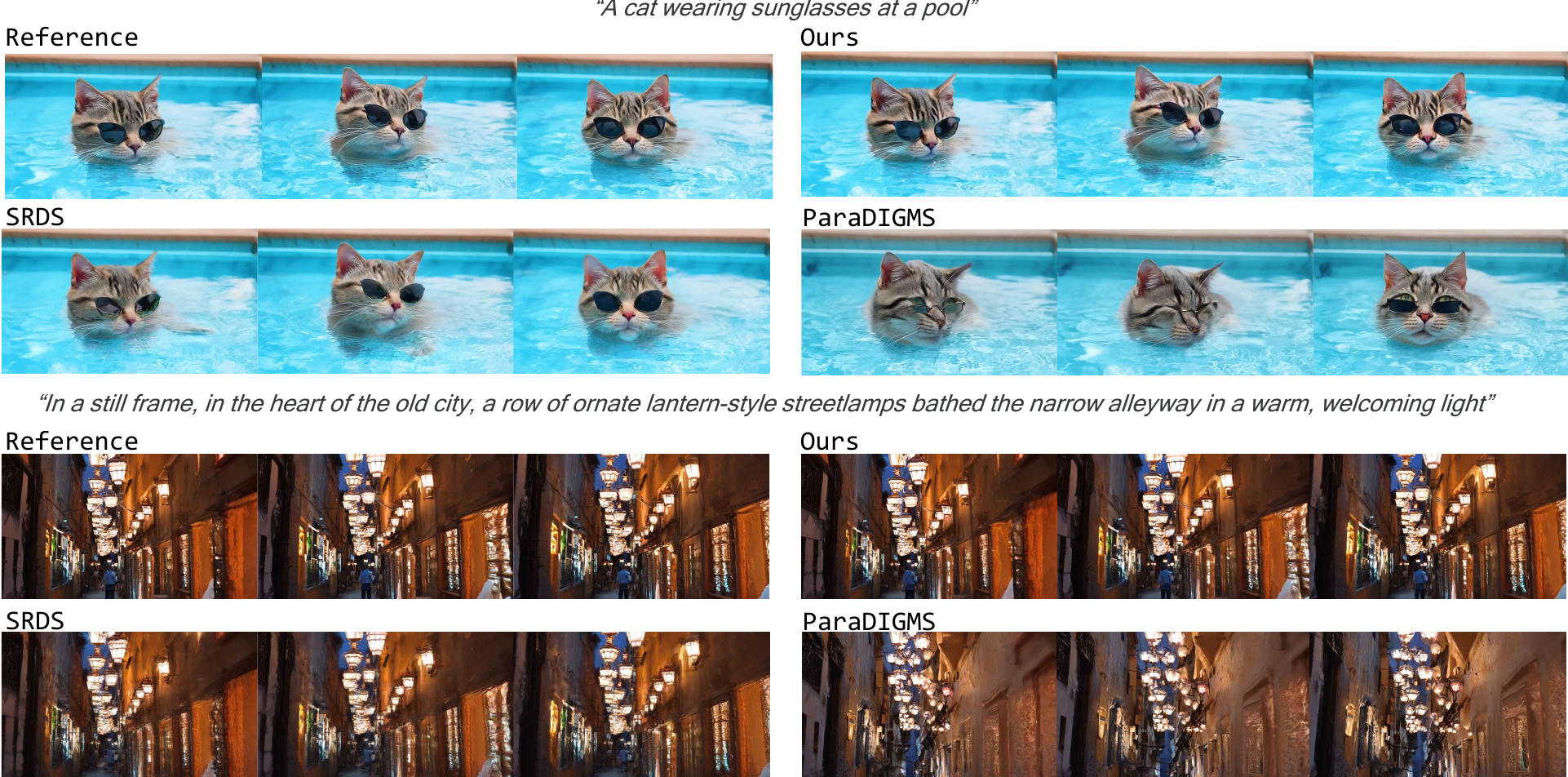}
    \vskip -0.1in
    \caption{Qualitative comparisons with baselines on CogVideoX1.5. Our approach achieves the best quality and the lowest latency.}
    \label{fig:video_comparison}
\end{figure*}

\begin{table*}[t!]
  \centering
  \caption{Benchmark results of parallel diffusion solvers on latent image diffusion models. We evaluate two models with 1000 prompts from the COCO2017 captions dataset. Our approach achieves the highest speedup without measurable quality degradation.}
  \vskip -0.1in
    \resizebox{\linewidth}{!}{%
    \begin{tabular}{clcccccccccccc}
    \toprule
          &       & \multicolumn{4}{c}{\textbf{Num Core = 4}} & \multicolumn{4}{c}{\textbf{Num Core = 6}} & \multicolumn{4}{c}{\textbf{Num Core = 8}} \\
          \cmidrule(lr){3-6}\cmidrule(lr){7-10}\cmidrule(lr){11-14}
          &       & Time per & Speed- & CLIP & Latent  & Time per & Speed- & CLIP & Latent  & Time per & Speed- & CLIP & Latent \\
          &       & sample $\downarrow$ & up $\uparrow$    & Score $\uparrow$ & RMSE $\downarrow$  & sample $\downarrow$ & up $\uparrow$    & Score $\uparrow$ & RMSE $\downarrow$  & sample $\downarrow$ & up $\uparrow$    & Score $\uparrow$ & RMSE $\downarrow$ \\
    \midrule
    \multirow{4}[2]{*}{SD-3.5-Large} & Sequential &  12.1     &   -    &   32.5    &  -     &   12.1    &     -  &   32.5    &   -    &   12.1    &   -    &    32.5   & - \\
          & ParaDIGMS~\cite{shih2024parallel} &  8.8     &   1.4    &  32.4     &  0.347     &  7.6     &  1.6     &   32.5    &   0.377    &  7.1     &   1.7    &   32.4    & 0.405 \\
          & SRDS~\cite{selvam2024selfrefining}  &  10.2     &  1.2     &     32.5    &    0.216   &  6.8     & 1.8      &    32.5     &   0.216    &    5.8   &   2.1    &  32.5     & 0.216 \\
          & \cellcolor{gray!15}Ours  &  \cellcolor{gray!15}{6.5}     &    \cellcolor{gray!15}{\textbf{2.0}}   &   \cellcolor{gray!15}{32.4}    &  \cellcolor{gray!15}{0.208}     &   \cellcolor{gray!15}{5.8}    &   \cellcolor{gray!15}{\textbf{2.2}}    &  \cellcolor{gray!15}{32.5}     &  \cellcolor{gray!15}{0.210}     &    \cellcolor{gray!15}{5.2}   &   \cellcolor{gray!15}{\textbf{2.4}}    &   \cellcolor{gray!15}{32.5}    & \cellcolor{gray!15}{0.211} \\
    \midrule
    \multirow{4}[1]{*}{Flux} & Sequential &   12.2    &   -    &  31.0     &   -    &   12.2    &   -    &   31.0    &  -     &   12.2    &   -    &   31.0    &  -\\
          & ParaDIGMS~\cite{shih2024parallel} &   8.6    &   1.4    &  31.0     &  0.223     &   8.1    &  1.5     &   31.1    & 0.256      &   7.5    &  1.6     &  31.0     & 0.301  \\
          & SRDS~\cite{selvam2024selfrefining}  &    9.6   &    1.3   &  31.0     &   0.183    &  6.4     &    1.9   &   31.0    & 0.183      &   5.5    &   2.3    &   31.0    &  0.183 \\
          & \cellcolor{gray!15}Ours  &  \cellcolor{gray!15}{6.6}     &    \cellcolor{gray!15}{\textbf{2.0}}   &   \cellcolor{gray!15}{31.1}    &  \cellcolor{gray!15}{0.173}     &   \cellcolor{gray!15}{5.8}    &   \cellcolor{gray!15}{\textbf{2.3}}    &  \cellcolor{gray!15}{31.0}     &  \cellcolor{gray!15}{0.178}     &    \cellcolor{gray!15}{4.9}   &   \cellcolor{gray!15}{\textbf{2.6}}    &   \cellcolor{gray!15}{31.0}    & \cellcolor{gray!15}{0.179} \\
    \bottomrule
    \end{tabular}%
    }
    \vskip -0.1in
  \label{tab:image_overall_table}%
\end{table*}%

\section{Experiments}\label{sec:exp}
In this section, we evaluate the empirical speedup of \alg~compared with existing parallel diffusion acceleration algorithms~\cite{shih2024parallel, selvam2024selfrefining} and the sequential solver. We consider three state-of-the-art video diffusion models and two image diffusion models. We also verify the efficacy of our core designs of~\alg~as well as its robustness to the allocated computation cores in~\Cref{sec:abl}. 

\subsection{Experimental Setups}


\noindent \textbf{Models.} For video generation, we benchmark our approach on three state-of-the-art large video diffusion models: HunyuanVideo~\cite{kong2024hunyuanvideo}, Wan2.1~\cite{wan2.1}, and CogVideoX1.5-5B~\cite{yang2024cogvideox}. For each model, we generate videos with prompts in VBench~\cite{huang2023vbench} strictly following VBench evaluation protocol.
We consider two powerful image diffusion models for image generation, Stable Diffusion 3.5 Large~\cite{esser2024scaling} and Flux~\cite{flux2024}, as the backbone. 
Following previous works~\cite{shih2024parallel,selvam2024selfrefining}, we sample 1000 prompts from COCO2017 captions dataset as the test bed.
We use $N=50$ diffusion steps by default, with more investigations on $N$ in~\Cref{sec:abl}.

%


\noindent \textbf{Methods.} We compare \alg~with the following baselines with 4, 6, and 8 cores, respectively. \emph{Sequential} is the sequential solver that iterates over $N$ steps, whose quality serves as the golden standard for parallel solvers. Following conventional choice, we adopt DDIM~\cite{song2020denoising} for diffusion and Euler~\cite{esser2024scaling} for flow matching. \emph{ParaDIGMS}~\cite{shih2024parallel} is a parallel diffusion sampler based on sliding-window Picard iteration, where the window size is the amount of cores under their setting; \emph{SRDS}~\cite{selvam2024selfrefining} is a recently proposed multi-grid parallel diffusion solver that enforces utilizing $\lfloor\sqrt{N}\rfloor$ number of cores. 
Since their method can be regarded as a special case in our framework, we additionally pipeline their algorithm to be more efficient for different $K$ (rather than a fixed number $\lfloor\sqrt{50}\rfloor$.
For \alg, the initialization sequence $\hat{\mathbf I}$ is set to $[0,8,16,32], [0,3,6,12,24,36],[0,2,4,8,16,24,32,40]$ for $K=4,6,8$, respectively.

\noindent \textbf{Metrics.} For both video and image models, we report \emph{Time per sample} that refers to the average wall-clock time used to generate one sample; \emph{Speedup} that refers to the relative speedup compared with sequential solve, measured by the number of sequential network forward calls. Notice that this will be slightly different from the measurement or the wall-clock. 
In terms of generation quality, we report
\emph{VBench Quality Score} as the standard quality metric obtained following the VBench evaluation protocol~\cite{huang2023vbench} for video generation, and \emph{CLIP Score}~\cite{hessel2021clipscore} evaluated using ViT-g-14~\cite{radford2021learning,ilharco_gabriel_2021_5143773} for the image generation.
We also report \emph{Latent RMSE} under both cases that measures the Rooted MSE between the returned latent of the algorithm and that of the sequential solver. Notice that a lower latent RMSE indicates lower sampling error, with sequential solve being the oracle.

\subsection{Accelerating Image and Video Diffusion}

\noindent \textbf{Video diffusion.} The benchmark results are presented in \Cref{tab:video_overall_table} with qualitative comparisons displayed in~\Cref{fig:video_comparison}. Our approach achieves remarkable speedups ranging from 2.4$\times$ to 2.9$\times$ across three large video diffusion models using 8 cores, without measurable quality degradation (unlike ParaDIGMS that suffers from much higher latent RMSE). When using only four cores, \alg~still offers significant speedup of ~2$\times$, while the latency of SRDS increases substantially (only~$1.3\times$ speedup on average).
Notice that the speedup of SRDS reported in~\Cref{tab:video_overall_table} is achieved with our unified pipeline, without such careful pipelining their speed could be even slower than Sequential.
This comparison further strengthens the broad applicability of our method with different compute resource budgets. 


\noindent \textbf{Image diffusion.} The benchmark results of image generation are presented in \Cref{tab:image_overall_table}. Similar to video generation, \alg~maintains significant speedups across different numbers of cores on image diffusion models, outperforming existing baselines by ~60\% with four cores and achieving ~$2.5\times$ speedup with eight cores.
Notice that this is obtained with the lowest latent RMSE and negligible change in CLIP Score, suggesting the superiority of \alg.

\subsection{Ablation Study and Analysis}
\label{sec:abl}

\begin{table}[!h]
  \centering
  \small
  \caption{Ablation study on the choice of the initialization sequence with video diffusion HunyuanVideo and image diffusion Flux.}
  \vskip -0.1in
  \setlength{\tabcolsep}{6.2pt}
        \resizebox{\linewidth}{!}{%
    \begin{tabular}{cccccc}
    \toprule
          &       & \multicolumn{2}{c}{\textbf{HunyuanVideo}} & \multicolumn{2}{c}{\textbf{Flux}} \\
           \cmidrule(lr){3-4}\cmidrule(lr){5-6}
    \multicolumn{1}{c}{Num} & \multicolumn{1}{c}{Initialization} & Speed- & VBench & Speed- & CLIP \\
    \multicolumn{1}{c}{Core} & \multicolumn{1}{c}{Sequence} & up $\uparrow$    & Quality $\uparrow$ & up $\uparrow$    & Score $\uparrow$ \\
    \midrule
      8    & Ours    &  \textbf{2.9}     &   84.1\%    &   \textbf{2.4}    &  31.0 \\
      8    & Uniform    &   2.6    &   84.0\%    &   2.2    &  30.9 \\
    \midrule
       6   & Ours     &  \textbf{2.5}     &   84.2\%    &  \textbf{2.3}     & 31.0 \\
      6    & Uniform     &   2.1    &   84.2\%    &   2.0    & 31.0 \\
    \midrule
       4   & Ours     &   \textbf{2.1}    &  84.1\%     &  \textbf{2.0}     & 31.1 \\
       4   & Uniform     &    1.8   &   84.2\%    &  1.8     &  31.0\\
    \bottomrule
    \end{tabular}%
    }
  \label{tab:ablation_initialization}%
\end{table}%

\begin{table}[!h]
  \centering
  \small
  \caption{The performance of \alg~on HunyuanVideo using eight cores, with different total diffusion sampling steps $N$.}
  \vskip -0.1in
  \setlength{\tabcolsep}{6.2pt}
    \begin{tabular}{ccccc}
    \toprule
       Total & Time per & Speed- & VBench & Latent \\
    Steps &  sample $\downarrow$    & up $\uparrow$ & Quality $\uparrow$    & RMSE $\downarrow$ \\
    \midrule
     50 &129.1 &2.9 & 84.1\%   & 0.068\\
     75& 165.8 & 3.4 & 84.4\% &  0.073 \\
     100 & 213.2 & 3.6 &  84.6\% & 0.076 \\
    \bottomrule
    \end{tabular}%
  \label{tab:different_N}%
  \vskip -0.1in
\end{table}%

\noindent \textbf{Design of initialization sequence $\hat{\mathbf I}$.} 
Framework~\ref{def:alg} indicates the optimal sequence $\mathbf I$ in theory, which inspires our choice of $\hat{\mathbf I}$. To empirically verify the necessity of such a design, we compare the performance with a uniform sequence (e.g., $\hat{\mathbf I} = [0, 6,12,18,24,30,36,42]$ for $K=8$), as shown in \Cref{tab:ablation_initialization}. We also provide the comparison on the convergence curve in the appendix. Clearly, the calibrated sequence consistently outperforms the uniform one across all evaluated scenarios, highlighting our theoretically derived strategy's effectiveness and practical value. Intuitively, our proposed sequence allocates shorter intervals to slower solvers, enabling more frequent and precise rectification at earlier stages. Consequently, this high-quality information is progressively propagated to faster solvers, substantially reducing errors and boosting the sample quality.

\begin{figure}[t!]
    \centering
    \includegraphics[width=\linewidth]{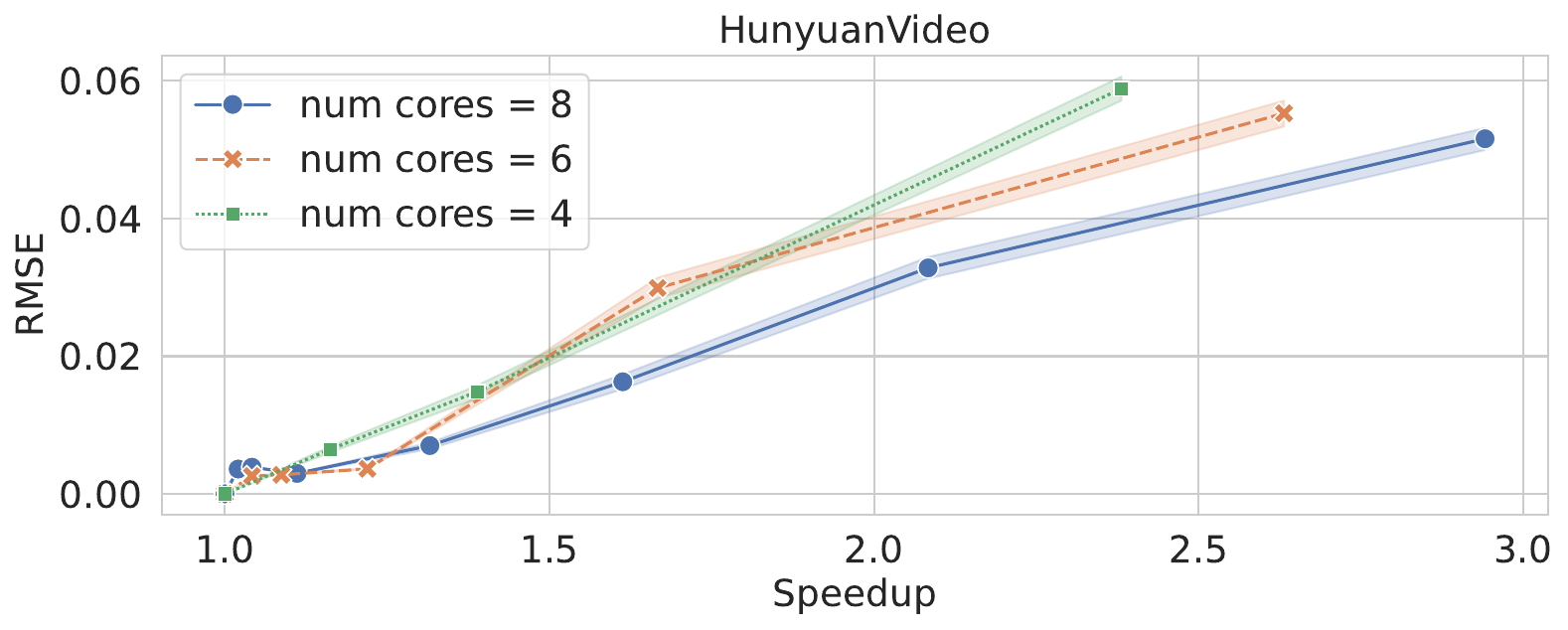}\\
    \includegraphics[width=\linewidth]{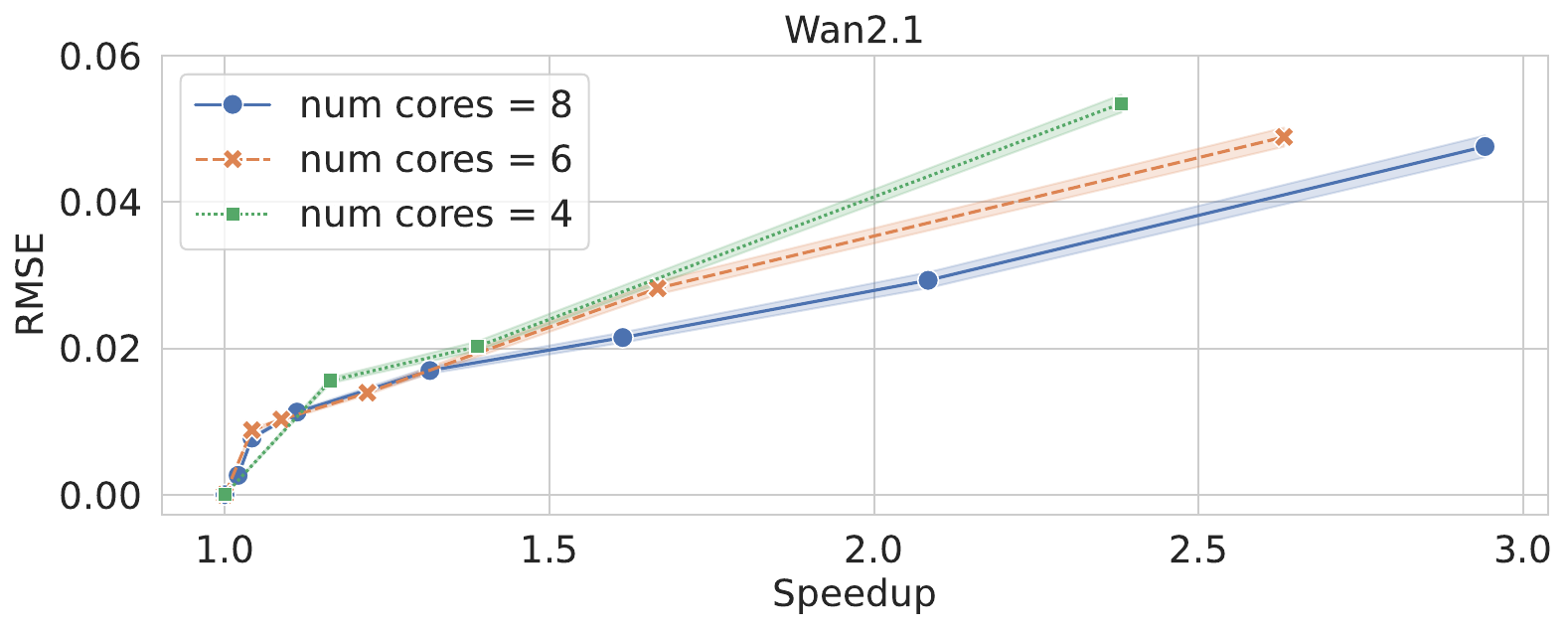}
    \vskip-0.1in
    \caption{Scaling of~\alg~towards different numbers of cores.}
    \label{fig:num_cores}
    \vspace{-10pt}
\end{figure}

\noindent \textbf{The effect of the number of cores $K$.} We investigate the scaling of~\alg~towards different numbers of cores with results in~\Cref{fig:num_cores}. We observe that our approach can benefit from additional parallel computing with improved empirical convergence achieved with a larger number of cores, and remains robust in cases with limited numbers of cores.

\noindent \textbf{Performance with different diffusion sampling steps $N$.} We additionally evaluate~\alg~towards different numbers of diffusion steps. The results in~\Cref{tab:different_N} demonstrate that~\alg~yields a consistent increase in speedup with larger numbers of diffusion steps while obtaining higher video quality due to fine-grained discretization.

\section{Related Works and Discussions}
\label{sec:discuss}

\noindent \textbf{Diffusion sampling acceleration.}
Significant progress has been made toward improving the latency of diffusion sampling, primarily via training-based distillation. This includes progressive distillation~\cite{salimans2022progressive}, consistency distillation~\cite{song2023consistency,kim2023consistency,song2024improved,lu2025simplifying}, and distribution-matching distillation~\cite{yin2024one,xie2024em,yin2024improved}, which compresses iterative procedures into fewer inference steps through specialized training. 
Alternatively, fast single-core ODE solvers~\cite{lu2022dpm,lu2022dpmpp,zhang2022fast,zhang2023gddim} aim at speeding up sampling without retraining by improving numerical integration. Without additional computation resources, these methods inevitably encounter a stringent trade-off between sampling speed and quality~\cite{zhang2023gddim}.

\noindent \textbf{Parallel sampling for diffusion models.}
Parallel computing provides another promising approach to accelerate diffusion by distributing sampling computations across multiple cores. Existing parallel sampling methods include Picard iteration~\cite{shih2024parallel}, Anderson acceleration~\cite{tang2024acceleratingparallelsamplingdiffusion}, self-refining diffusion sampling~\cite{selvam2024selfrefining}, and 3D decomposition methods~\cite{zhou2024dreampropeller}. However, these approaches either require pre-defined convergence thresholds, heavily depend on heuristics, or impose inflexible resource allocation constraints due to fixed initialization. Another orthogonal line exploits parallelism within neural network operations, such as distributing attention computations across GPUs~\cite{fang2024xdit}. Remarkably, this line of methods can be seamlessly combined with \alg~to achieve complementary acceleration, pushing the limits toward real-time generation.

\noindent \textbf{Diffusion streaming.}
Our hierarchical multi-core solver naturally supports a ``diffusion streaming'' paradigm by progressively refining output quality as the sampling proceeds across cores. This streaming capability empowers users to flexibly balance generation speed and sample quality by selecting intermediate outputs on the fly, setting our method apart from existing parallel algorithms that rely heavily on pre-defined convergence thresholds or stopping criteria~\cite{shih2024parallel}. Consequently, our approach is highly suitable for interactive, latency-sensitive deployment scenarios.




\section{Conclusion}

This paper introduces \alg, a novel multi-core hierarchical ODE-based algorithm for multi-core diffusion acceleration by progressively rectifying the faster core's latents by the slower core's accurate solutions.
Empirically, \alg~demonstrates substantial acceleration across diverse architectures and modalities, achieving remarkable inference-time reductions without sacrificing generation quality. \alg~presents a significant step toward real-time diffusion sampling in latency-sensitive applications.

\clearpage

\noindent \textbf{Acknowledgments.}
We thank the anonymous reviewers for their  feedback on improving the manuscript. We thank Amil Merchant and Nikil Roashan Selvam for helpful discussions. This work was supported by ARO
(W911NF-21-1-0125), ONR (N00014-23-1-2159), and the CZ Biohub.

{
    \small
    \bibliographystyle{ieeenat_fullname}
    \bibliography{main}
}

\appendix
\newpage

\onecolumn
\section{Proofs}
\setlength{\parskip}{5pt}
\subsection{Proof of Lemma \ref{lemma:rectification}}
\begin{proof}


Since \(\mathbf f\) is sufficiently smooth in a neighborhood of \((\rvx_t,t)\), near \(t\), we can expand:

\[
\rvx_{t'} 
= 
\rvx_t
\;+\; 
(t' -t)\,\mathbf f\bigl(\rvx_t,t\bigr)
\;+\;
\frac{(t' - t)^2}{2}\,\Bigl[\partial_t \mathbf f(\rvx_t,t) + \nabla_\rvx \mathbf f(\rvx_t,t)\cdot \mathbf f(\rvx_t,t)\Bigr]
\;+\;
O\bigl((t' - t)^3\bigr),
\]
and
\[
\tilde{\rvx}_{t'} 
= 
\tilde{\rvx}_t
\;+\; 
(t' - t)\, \mathbf f\bigl(\tilde{\rvx}_t,t\bigr)
\;+\;
\frac{(t' - t)^2}{2}\,\Bigl[\partial_t \mathbf f(\tilde{\rvx}_t,t) + \nabla_\rvx \mathbf f(\tilde{\rvx}_t,t)\cdot \mathbf f(\tilde{\rvx}_t,t)\Bigr]
\;+\;
O\bigl((t' - t)^3\bigr).
\]

Subtracting the expansions, we get
\[
\begin{aligned}
\rvx_{t'} - \tilde{\rvx}_{t'}
&=\;
\Bigl[\rvx_t - \tilde{\rvx}_t\Bigr]
\;+\;
(t' - t)\,\Bigl(\mathbf f(\rvx_t,t) - \mathbf f(\tilde{\rvx}_t,t)\Bigr)
\\[6pt]
&\quad+\;
\frac{(t' - t)^2}{2}\,\Bigl\{\partial_t \mathbf f(\rvx_t,t)+ \nabla_\rvx \mathbf f(\rvx_t,t)\cdot \mathbf f(\rvx_t,t)
\;-\;
\bigl[\partial_t \mathbf f(\tilde{\rvx}_t,t) + \nabla_\rvx  \mathbf f(\tilde{\rvx}_t,t)\cdot \mathbf f(\tilde{\rvx}_t,t)\bigr]\Bigr\}
\;+\;
O\bigl((t' - t)^3\bigr).
\end{aligned}
\]
Hence for small \(t' - t\),
\[
\rvx_{t'} - \tilde{\rvx}_{t'}
= 
\bigl[\rvx_t - \tilde{\rvx}_t\bigr] 
\;+\;
(t' - t)\,\Bigl[\mathbf f(\rvx_t, t) - \mathbf f(\tilde{\rvx}_t,t)\Bigr] 
\;+\;
O\bigl((t' - t)^2\bigr).
\]

After the rectification
\[
\hat{\rvx}_{t'} 
=  \tilde \rvx_{t'} + \mathbf r_\theta (\rvx_t, \tilde \rvx_t, t, t'-t) =
\tilde{\rvx}_{t'} 
\;+\;
(t' - t) \,\Bigl[\mathbf f(\rvx_t,t) - \mathbf f(\tilde{\rvx}_t,t)\Bigr] + \bigl[\rvx_t - \tilde{\rvx}_t\bigr] .
\]

Then
\[
\begin{aligned}
\rvx_{t'} - \hat{\rvx}_{t'}
= &
\bigl[\rvx_{t'} - \tilde{\rvx}_{t'}\bigr]
\;-\;
(t' - t)\,\bigl[\mathbf f(\rvx_t,t) - \mathbf f(\tilde{\rvx}_t,t)\bigr] - \bigl[\rvx_t - \tilde{\rvx}_t\bigr]\\
= & O\bigl((t' - t)^2\bigr).
\end{aligned}
\]
Since \(\rvx_{t'} - \tilde{\rvx}_{t'}\) has precisely a linear term in \(t' - t\) given by 
\(\,(t' - t)\,[\mathbf f(\rvx_t,t) - \mathbf f(\tilde{\rvx}_t,t)]\,\) plus possibly lower-order terms, we can conclude that 
\[
\|\rvx_{t'} - \hat{\rvx}_{t'} \|_2  = o(
\|\rvx_{t'} - \tilde{\rvx}_{t'}\|_2),\]
which completes the proof.


\end{proof}






    

\subsection{Proof in Definition \ref{def:reward}}

Throughout the proof in this and the next section, we show the result for each coordinate of $\rvx$, and the claim holds by aggregating all the coordinates together. We use $x$ to represent a coordinate of $\rvx$.
\begin{proof}
    We prove each property separately.

    Given that $\rvf_\theta(x, t) = x $ and $x_0 = 1$, it can be easily derived that $x_t = e^t$.

    \textbf{Optimality.} For single-core solve, $\mathcal{R}([0]) = \ln e^1 = 1$, and $\mathcal{S}([0]) = \frac{1}{1 - t_0} = \frac{1}{1 - 0} = 1$. For any $\mathbf{I}$ such that $\mathcal{S}(\mathbf I) > 1$, we have $t^K > 0 $. Denote $x^{i}_{t}$ as the solution of the $i$-th core at time $t$. Then we have $x^{1}_{t} = 1 + t$, $x^{0}_{t} = e^t$. Since $e^t > 1 + t$, we have $x^{0}_{t} > x^{1}_{t}$, for any $t > 0$. Suppose we have $x^{i}_{t} > x^{i+1}_{t}$, then $x^{i+1}_{t'} = (e^{t' - t} - (t' - t) - 1) x^{i+1}_{t} + ((t' - t) + 1) x^{i}_{t} < e^{t'-t}x^{i}_{t} \leq x^{i}_{t'}$ for some $t < t'$. Hence, by induction we have $x^{i}_{t} > x^{i+1}_{t}$ for any $i \leq K - 1, t \in [0, 1]$. In that way, we can calculate that $\mathcal R (\mathbf I) < \ln x^{0}_{1 - t^{(K)}} = \ln e^{1 - t^{(K)}} = 1 - t^{(K)} < 1$.

    \textbf{Monotonicity.} By induction, it suffices to show that inserting a core to $\mathbf{I}_1$ while not changing $\mathcal{S}(\mathbf I_1)$ will improve the reward $\mathcal{R}(\mathbf{I}_1)$. By the derivation in the part of optimality, and by Lemma \ref{lemma:rectification}, we have that updating with rectification can increase the value at the same time step. Suppose the solution at time $t$ on the new inserted core is $x^{\mathrm{new}}_{t}$, and suppose the new core is inserted between core $i$ and core $i + 1$. We now prove that $x^{i+1}_{t}$ increases for any $t>0$. For $t_1 < t$, we calculate the contribution of $x^{i}_{ t_1}$ to $x^{i+1}_{t}$ through rectification. Here, we choose $t_1$ to be the smallest $t_1$ such that $x^{i}_{ t_1}$ has a contribution to the current rectification but has no contribution to the previous rectification. Before inserting $x^{\mathrm{new}}$, the contribution of $x^{i}_{ t_1}$ to $x^{i+1}_{t}$ through rectification is $x^{i}_{ t_1} (1+t - t_1)e^{t - t_1}$. After inserting $x^{\mathrm{new}}$, this contribution will either remain the same, or increase due to the new rectification interacting with $x^{\mathrm{new}}$. Specifically, if rectification happens with $x^{\mathrm{new}}$, then this contribution will be $(1+s_1)(1+s_2)e^{t-t_1}$. Here, $s_1,s_2>0$ are two times such that $s_1+s_2 = t-t_1$, thus $(1+s_1)(1+s_2) = 1+t-t_1+s_1s_2 > 1+ t-t_1$. This is because the accurate solver will always contribute $e^{t-t_1}$, while the coarse solver will be improved by adding a middle point. Notice that there must be at least one rectification happening with $x^{\mathrm{new}}$, $x^{i+1}_{t}$ will increase as a result.
    
    Then, by induction, it can be easily shown that $\mathcal{R}(\mathbf{I}_1 )$ increases after core insertion, which indicates, again by induction, $\mathcal{R}(\mathbf{I}_1) < \mathcal{R}(\mathbf {I}_2) $ for $\mathbf I_1 \subsetneq \mathbf I_2$. 

    For any $\mathbf{I}_1$ as the prefix of $\mathbf{I}_2$, suppose the last time step of $\mathbf{I}_1$ is $t_{K_1}$ and the last time step of $\mathbf{I}_2$ is $t_{K_2}$, with $K_1 \leq K_2$. From the result in the proof of optimality, we have that $x^{K_1}_{1} \geq x^{K_2}_{1}$, which means $\mathcal{R}(\mathbf{I}_1) \geq \mathcal{R}(\mathbf{I}_2)$.

    \textbf{Trade-off.} For any speed-up ratio $1 \leq s_1 < s_2$, suppose by contradiction we have $\max_{\mathcal S(\mathbf I) = s_1} \mathcal R(\mathbf I) \leq \max_{\mathcal S(\mathbf I) = s_2} \mathcal R(\mathbf I)$. In this case, there exists $I_2$ such that $\mathcal{S}(\mathbf{I}_2) = s_2$ and for any $\mathbf{I}_1$ with $\mathcal{S}(\mathbf{I}_1) = s_1 $ we have $\mathcal{R}(\mathbf{I}_2) \geq \mathcal{R}(\mathbf{I}_1)$. Still by the result in monotonicity, we can insert an additional core to $\mathbf{I}_2$ to be $\mathbf{I}_2'$ such that $\mathbf{I}_2 \subset \mathbf{I}_2'$ and $1 - \frac{1}{s_1} \in \mathbf{I}_2'$. Further, $\mathcal{R}(\mathbf{I}_2') \geq \mathcal{R}(\mathbf{I}_2)$. However, based on the part of monotonicity, we can find $I_1'$ that is a prefix of $\mathbf{I}_2'$ and $\mathcal{S}(\mathbf{I}_1') = s_1$, and $\mathcal{R}(\mathbf{I_1'}) > \mathcal{R}(\mathbf{I}_2')$, which is a contradiction. By contradiction, we completed the proof.
    
\end{proof}

\subsection{Proof of Theorem \ref{thm:core_insert}}
\begin{proof}

  Denote $x^{i}_{t}$ as the solution of the $i$-th core at time t, $i=1,2,3$.
    Notice that the update solver solving $x^{i}_{t} \rightarrow x^{j}_{ t'}$ follows the following rules: (1) Fine solver: $F(x^{i}_{t} \rightarrow x^{i}_{t'}) = x^{i}_{t} e^{t' - t}$; (2) Coarse solver: $G(x^{i}_{t} \rightarrow x^{j}_{ t'}) = x^{i}_{t} (1 + t' - t)$, for $i = j$ or $(i,j) = (1,2), (2,3)$.

    Suppose $T = [0, t, \frac{s - 1}{s}]$. First, we prove the case for $s \leq 3$. 
    
    If $t < 1 - \frac{2}{s}$, then the second core will not help update any point on the trajectory of the third core, which is trivial, so w.l.o.g we assume $t \geq 1 - \frac{2}{s}$.

    \textbf{Case 1.} $s\leq 3$, $1 - \frac{2}{s} \leq t \leq \frac{s-1}{2s}$. 
    
    Since $s > 2$, we have $1 - \frac{1}{s} - t \geq 1 - \frac{1}{s} - \frac{s-1}{2s} > \frac{1}{2s}$, which means in this case the trajectory of the third core will have only one multi-core communication update. Suppose there are $k - 1$ communications between the first and the second cores, then w.l.o.g we can assume $t = (1 - \frac{1}{s})\frac{1}{k}$.

Now, by update rules, we have $x^{1}_{t} = e^{t}$, $x^{2}_{ t} = 1 + t$, 

$$
\begin{aligned}
    x^{2}_{ 2t} = &F(x^{2}_{ t}\rightarrow x^{2}_{ 2t}) + G(x^{2}_{ t} \rightarrow x^{2}_{2t}) - G(x^{1}_{2t} \rightarrow x^{2}_{t}), \\
    = &e^t(1+t) + (1+t)(e^t - t -1),\\
    = & e^{2t} - (e^t - t - 1)^2.
\end{aligned}
$$

By induction, we can show that for $i=1,\ldots, k$
$$
\begin{aligned}
    x^{2}_{ it} = &F(x^{2}_{ (i-1)t)}\rightarrow x^{2}_{ it}) + G(x^{2}_{ (i-1)t} \rightarrow x^{2}_{ it}) - G(x^{1}_{ (i-1)t} \rightarrow x^{2}_{ it}),\\
    = & (e^t - t - 1)(e^{(i-1)t} - (e^t - t - 1)^{i-1}) + (1 + t)e^{(i-1)t},\\
    = & e^{it} - (e^t - t - 1)^i.
\end{aligned}
$$

Hence, $x^{2}_{ kt} = e^{kt} - (e^t - t - 1)^k$. By the update rule $x^{3}_{ 1 - \frac{1}{s} + (k-1)t} = F(x^{3}_{ 1 - \frac{1}{s}} \rightarrow x^{3}_{ 1 - \frac{1}{s} + (k-1)t}) + G(x^{3}_{ 1 - \frac{1}{s}} \rightarrow x^{3}_{ 1 - \frac{1}{s} + (k-1)t}) - G(x^{2}_{ 1 - \frac{1}{s}} \rightarrow x^{3}_{ 1 - \frac{1}{s} + (k-1)t})$, and $x^{3}_{ 1} = F(x^{3}_{ 2 - \frac{1}{s} + (k-1)t} \rightarrow x^{3}_{1})$, we can calculate that 

$$
\begin{aligned}
    x^{3}_{1} = &e^{1 - (2k-1)t}\left[ \left(e^{(k-1)t} - (k-1)t - 1\right)\left(1 + t\right)\left(1 + (k-1)t\right) + \left((k-1)t + 1\right)\left(e^{kt} - (e^t - t - 1)^k\right)  \right],\\
    = & e^{1 - (2k-1)t} (1+(k-1)t)\left[e^{kt} - (e^t - t - 1)^k + (1+t) (e^{(k-1)t} - (k-1)t - 1)\right].\\
\end{aligned}
$$
 It'll be shown later that $x^{3}_{ 1}$ is maximized at $k=2$, which corresponds to $t = \frac{s - 1}{2s}$. 

\textbf{Case 2.} $s \leq 3$, $\frac{s - 1}{2s} \leq t < \frac{s - 1}{s}$.

Suppose there are $k - 1$ communications between the second core and the third core, which gives $t = (1 - \frac{1}{s})\frac{k-1}{k}$. Since $c > 2$, for the same reason as in Case 1, there will be only one communication between the first core and the second core. Now, we have $x^{2}_{ (1 - \frac{1}{s})\frac{k-1}{k}} = 1 + (1 - \frac{1}{s})\frac{k-1}{k}$, $x^{3}_{ 1 - \frac{1}{s}} = (1 + (1 - \frac{1}{s})\frac{k-1}{k})(1 + (1 - \frac{1}{s})\frac{1}{k})$. Denote $\frac{1}{k}(1 - \frac{1}{s}) = a$, then $x^{2}_{ (1 - \frac{1}{s})\frac{k-1}{k}} = 1 + (k-1)a$, $x^{3}_{ 1 - \frac{1}{s}} = (1 + (k-1)a) (1 + a)$.  

Denote $f(i) = x^{3}_{ (k + i)a}$, then $f(0) =  (1 + (k-1)a) (1 + a)$. And further, we have 

\begin{equation}\label{update_rule}
    f(i) = (e^a - a - 1)f(i-1) + (a+1) (1 + (k-1)a)e^{ia},
\end{equation}
for $i=1, \ldots, k - 2$. This gives $\frac{f(i)}{(e^a - a - 1)^i} = \frac{f(i-1)}{(e^a - a - 1)^{i - 1}} + (1+a)(1+(k-1)a)\left(\frac{e^a}{e^a - a - 1}\right)^i$. By induction, we can calculate that $\frac{f(i)}{(e^a - a - 1)^i} = (1+a)(1+(k-1)a)\frac{\left(\frac{e^a}{e^a - a - 1}\right)^{i+1} - 1}{\frac{e^a}{e^a - a - 1} - 1}$, which gives 
\begin{equation}\label{update}
    f(i) = (1+a)(1+(k-1)a)\frac{e^{(i+1)a} - (e^a - a -1)^{i+1}}{1+a} = (1 + (k-1)a)(e^{(i+1)a} - (e^a - a -1)^{i+1}).
\end{equation}

 Now we get that $f(k-2) = (1+(k-1)a)(e^{(k-1)a} - (e^a - a -1)^{k-1})$. 

Since communication happens between the first and second core at time $2(k-1)a$, we have 

$$
\begin{aligned}
    f(k-1) = &(e^a - a - 1)f(k-2) + (1+a)(1 + (k-1)a)(2e^{(k-1)a} - (k-1)a - 1),\\
    = & (e^a - a - 1)(1+(k-1)a)(e^{(k-1)a} - (e^a - a -1)^{k-1}) + (1+a)(1 + (k-1)a)(2e^{(k-1)a} - (k-1)a - 1),\\
    = & (1 + (k-1)a)\left(e^{ka} - (e^a - a - 1)^k + (a+1)(e^{(k-1)a} - (k-1)a - 1)\right).
\end{aligned}
$$

Finally, we get 

$$
\begin{aligned}
    f(k) = &e^{1- (2k-1)a} f(k-1), \\
    = &e^{1-  (2k-1)a}(1 + (k-1)a)\left(e^{ka} - (e^a - a - 1)^k + (a+1)(e^{(k-1)a} - (k-1)a - 1)\right).
\end{aligned}
$$

Notice that $x^{3}_{1} = f(k)$ and this is exactly the same formula as in Case 1. Hence we have $x^{3}_{1}$ maximized at $k=2$, which corresponds to $t = \frac{s-1}{2s}$.

Next, we prove the case for  $s > 3$. In this case, the first core is completely idle in that its communication with the second core won't help the hit. To ensure that there exists at least one communication between the second and the third core, we should consider $t \geq 1 - \frac{2}{s}$. 

\textbf{Case 3.} $s \geq 3$, $1 - \frac{2}{s} \leq t \leq 1 - \frac{1}{s}$.

Suppose there are $k$ communications between the second and the third core. In this case, $t = 1 - \frac{1}{s}(1 + \frac{1}{k})$.

It can be calculated that $x^{2}_{ t} = 1 + t$, $x^{3}_{ 1 - \frac{1}{s}} = (1+t)(2- \frac{1}{s} - t)$. Denote $a = \frac{1}{sk}$. Notice that this is exactly the update of Case 2, except that there is no communication between the first core and the second core. Hence, by the update rule \ref{update_rule} and the derivation of \ref{update}, we can get that 

$$
x^{3}_{ 1} = (2 - (k+1)a)(e^{(k+1)a} - (e^a - a - 1)^{k+1}).
$$

Next, we prove that $x^{3}_{ 1}$ is maximized at $k=1$, which corresponds to $t = 1 - \frac{2}{s}$. The other two cases follow a similar argument.

Denote $c = \frac{1}{s}$, then $c \in (0, \frac{1}{3}]$. When $k=1$, $a = \frac{1}{s} = c$. Now it suffices to prove 

$$
(2 - c - a) (e^{a + c} - (e^a - a - 1)^{\frac{c}{a} + 1}) \leq (2 - 2c)(c+1)(2e^c - c - 1),
$$

which is equivalent to proving, for $a \in (0, c]$, 

\begin{equation}\label{eq:1}
(2 - c -a )(a+1)\frac{e^a}{a+1}(1 - (1 - \frac{a+1}{e^a})^{\frac{c}{a} + 1}) \leq (2 - 2c)(c+1) (2 - \frac{c+1}{e^c}).
\end{equation}

Notice that $(2 - c - a)(a+1) = -a^2+ (1-c)a + 2-c$ increases in $a$ for $a \leq c \leq \frac{1-c}{2}$. The last equality holds because $c \leq \frac{1}{c}$. Therefore, we have $(2 - c -a )(a+1) \leq (2 - 2c)(c+1)$. To show \eqref{eq:1}, it remains to show 

$$
\frac{e^a}{a+1}(1 - (1 - \frac{a+1}{e^a})^{\frac{c}{a} + 1}) \leq 2 - \frac{c+1}{e^c}.
$$

Define $h(c)= \frac{e^a}{a+1}(1 - (1 - \frac{a+1}{e^a})^{\frac{c}{a} + 1}) - 2 + \frac{c+1}{e^c}$, then it suffices to show $h(c) \leq 0$ for $c \geq a$. Since $h(a) = 0$, it then suffices to show $h'(c) \leq 0 $ for $c \geq a$.

$$
\begin{aligned}
    h'(c) = & \frac{e^a}{a+1 }(\frac{a+1}{e^a} - 1)(1 - \frac{a+1}{e^a})^{\frac{c}{a}}\ln(1-\frac{a+1}{e^a})\frac{1}{a} - \frac{c}{e^c},\\
    = & \frac{1}{e^c}\left((1 - \frac{e^a}{a+1})(e^a - a - 1)^{\frac{c}{a}}\ln(1 - \frac{a+1}{e^a})\frac{1}{a} - c\right).
\end{aligned}
$$

Now it suffices to show $(1 - \frac{e^a}{a+1})(e^a - a - 1)^{\frac{c}{a}}\ln(1 - \frac{a+1}{e^a})\frac{1}{a} - c \leq 0$.

Denote $g(c) = (1 - \frac{e^a}{a+1})(e^a - a - 1)^{\frac{c}{a}}\ln(1 - \frac{a+1}{e^a})\frac{1}{a} - c$, then we have 

$$
g'(c) = (1 - \frac{e^a}{a+1})(e^a - a - 1)^{\frac{c}{a}}\ln(1 - \frac{a+1}{e^a})\frac{1}{a^2}\ln(e^a - a - 1) - 1,
$$

$$
g''(c) = (1 - \frac{e^a}{a+1})(e^a - a - 1)^{\frac{c}{a}}\ln(1 - \frac{a+1}{e^a})\frac{1}{a^3}\left[\ln(e^a - a - 1)\right]^2 > 0,
$$

Since $g''(c) > 0$, $g'(c)$ monotonically increases in $c$. Next, we show $g'(c)\leq 0$, to show which suffices to show $\lim_{c \rightarrow \infty} g'(c) \leq 0$. Indeed, since $a \in (0, \frac{1}{3}]$, $e^a - a - 1 \leq e^{\frac{1}{3}} - \frac{4}{3} < 1$, hence $\lim_{c \rightarrow \infty} g'(c)) = 0 - 1 = -1 < 0$. Now we proved that $g'(c) \leq 0$ for $c \geq a$.

Finally, it remains to show $g(a) \leq 0$ for $a \in (0, \frac{1}{3}]$, which is 

$$
(1 - \frac{e^a}{a+1})(e^a - a - 1)\ln(1 - \frac{a+1}{e^a})\frac{1}{a} - a \leq 0.
$$

This is equivalent to proving 

$$
(e^a - a - 1)^2 \ln(1 - \frac{a+1}{e^a}) + a^2(a+1) \geq 0.
$$

By the inequality $\ln(1+x) \geq \frac{x}{\sqrt{1+x}}$ for $x \in (-1,0]$, we have 
$(e^a - a - 1)^2 \ln(1 - \frac{a+1}{e^a}) \geq (e^a - a - 1)^2 \frac{-\frac{a+1}{e^a}}{\sqrt{1 - \frac{a+1}{e^a}}} = \frac{(e^a - a - 1)^{\frac{3}{2}}}{e^{\frac{a}{2}}}(-a-1)$. Now it suffices to show $\frac{(e^a - a - 1)^{\frac{3}{2}}}{e^{\frac{a}{2}}} \leq a^2$, which is equivalent to $\frac{(e^a - a - 1)^3}{e^a} \leq a^4$.

Indeed, $\frac{(e^a - a - 1)^3}{e^a} \leq (e^a - a - 1)^2 \leq a^4$. The last inequality holds because $e^a - 1 - a - a^2 \leq 0$ for $a\in(0, \frac{1}{3}]$. Now we have completed the proof.






 


\end{proof}

\section{More Experiment Results}

\begin{figure}[t!]
    \centering
    \includegraphics[width=0.60\linewidth]{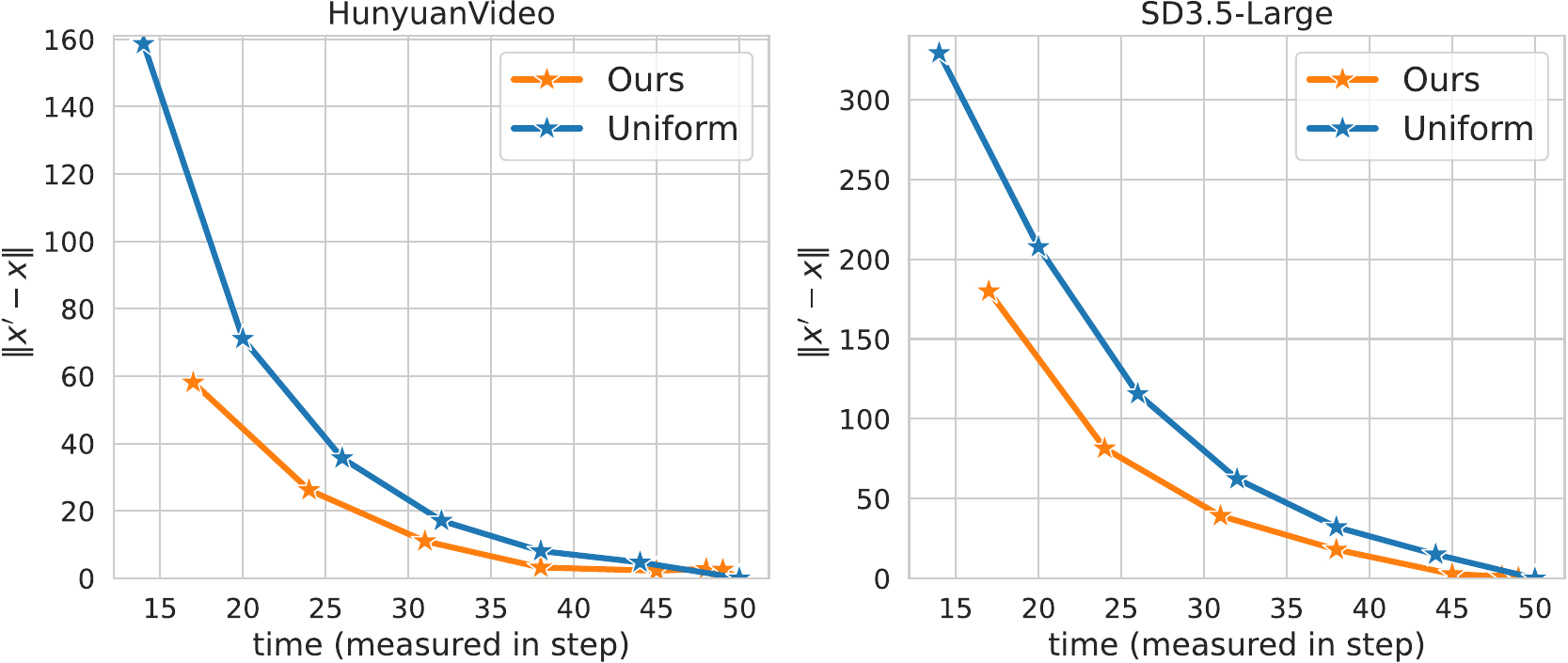}
    \caption{Convergence curve on HunyuanVideo and SD3.5-Large.}
    \label{fig:convergence}
\end{figure}

\noindent \textbf{Convergence curve of different initialization sequences.}
We additionally provide Figure~\ref{fig:convergence} to show the convergence curve of the L1-distance between the early-hit samples and the final output. The result strongly verifies the importance of initialization by showing that our strategy yields remarkably faster convergence compared with uniform initialization sequence.

\noindent \textbf{More qualitative results.} Here we provide more qualitative results of our model in different video diffusion models and image diffusion models. Note that we also present additional video samples obtained by our approach on the \textbf{project website}.

\begin{figure}[htbp]
    \centering
    \includegraphics[width=\linewidth]{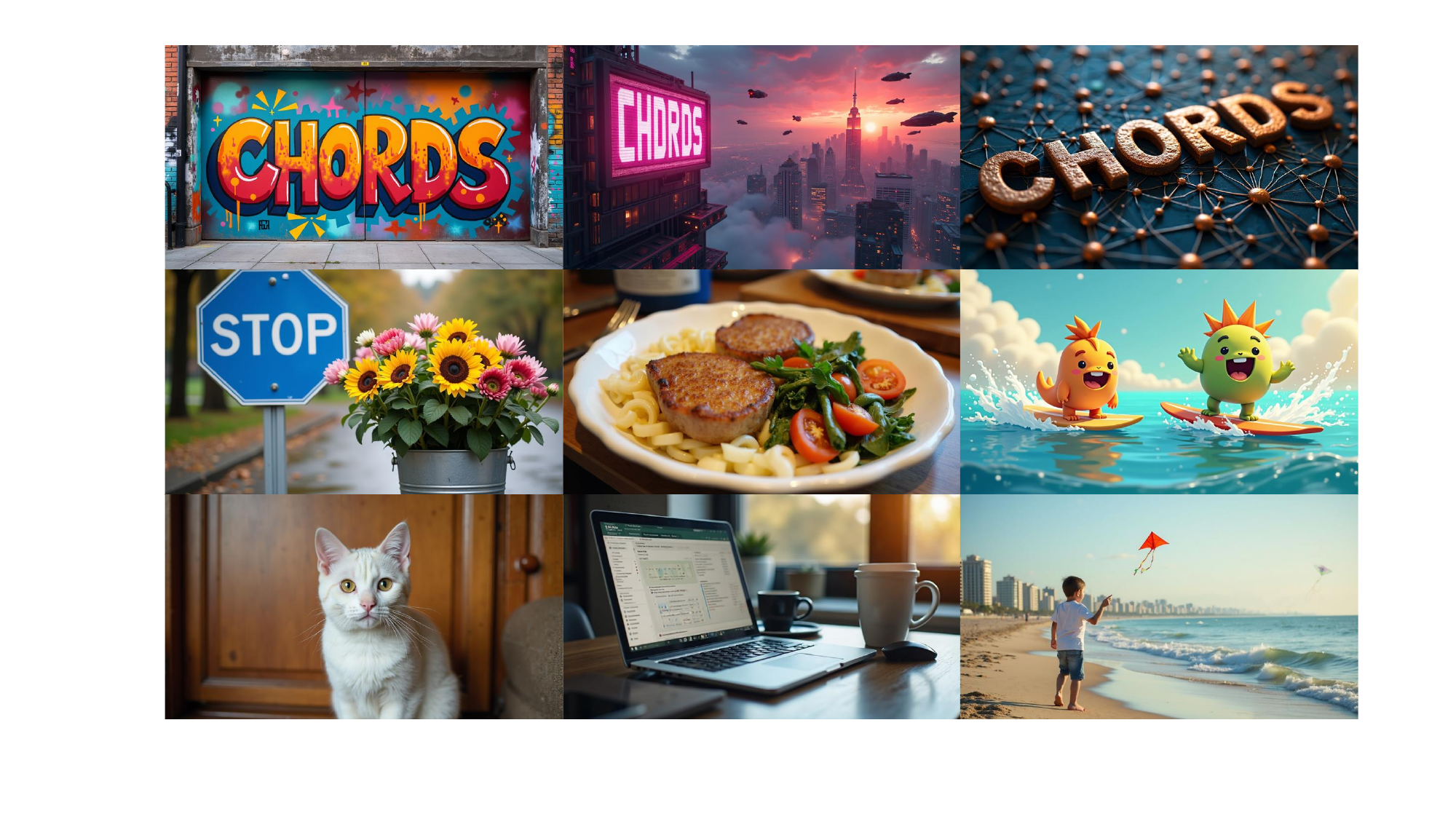}
    \caption{More results on image diffusion model Flux.}
    \label{fig:enter-label}
\end{figure}

\begin{figure}[htbp]
    \centering
    \includegraphics[width=\linewidth]{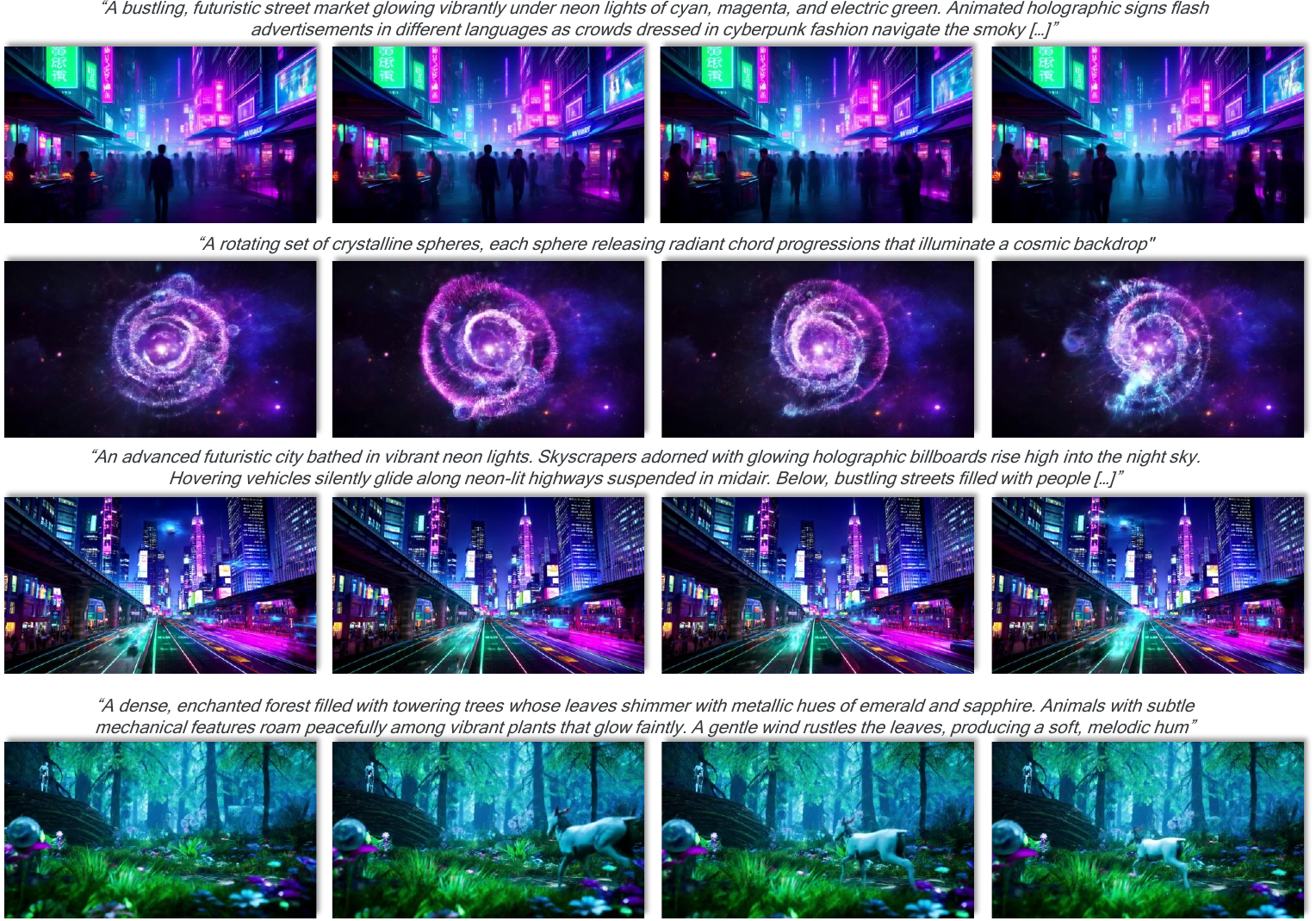}
    \caption{More results on video diffusion model HunyuanVideo.}
\end{figure}

\end{document}